\documentclass[journal,comsoc]{IEEEtran}
\usepackage[T1]{fontenc}% optional T1 font encoding
\usepackage{graphicx}
\usepackage{comment}
\usepackage{amsmath,amssymb} % define this before the line numbering.
\usepackage{color}
\usepackage{epsfig}
\usepackage{amsmath}
\usepackage{amssymb}
\usepackage{subfigure}
\usepackage{amssymb}
\usepackage{mathrsfs}
\usepackage{color}
\usepackage{float}
\usepackage{caption}
\usepackage{colortbl}
\usepackage{arydshln}
\usepackage{multirow}
\usepackage{multicol}
\usepackage[switch]{lineno}

\usepackage{hyperref}
\hypersetup{colorlinks,allcolors=black}
\definecolor{citecolor}{RGB}{34,139,34}
\newcommand{\etal}{\emph{et al. }}
\newcommand{\eg}{\emph{e.g. }}
\newcommand{\ie}{\emph{i.e. }}
\newcommand{\yx}[1]{\textcolor{black}{#1}}
\newcommand{\yxnew}[1]{\textcolor{black}{#1}}

\ifCLASSINFOpdf
\else
\fi
\usepackage{amsmath}
\usepackage[cmintegrals]{newtxmath}
\begin{document}
\title{Sketch-Guided Scenery Image Outpainting}
%\linenumbers

\author{
%\thanks{This work was supported in part by the NSFC under Grant 161772407, 61701391, 61732008, and 61902309 and ARC DECRA DE190101315, ARC DP200100938. This paper is also in part supported by Emergence Mechanism and Calculation Method of Group Intelligence based on Internet with No: 2018AAA0101100; National Key Research and Development Project with No: 2019YFB2102500. }
Yaxiong Wang,
        Yunchao Wei, Xueming Qian,~\IEEEmembership{Member~IEEE, } Li Zhu, and Yi Yang% <-this % stops a space
\IEEEcompsocitemizethanks{\IEEEcompsocthanksitem Y. Wang is with the School of Software Engineering, Xi'an Jiaotong University, Xi'an, 710049, China. He is now a visiting Ph.D student at ReLER Lab, University of Technology Sydney  \protect
 (e-mail: wangyx15@stu.xjtu.edu.cn).
\IEEEcompsocthanksitem Y. Wei is with the Centre for Artificial Intelligence, University of Technology Sydney, Ultimo, NSW 2007, Australia
 (e-mail: wychao1987@gmail.com, co-corresponding author).\protect% <-this % stops an unwanted space
\IEEEcompsocthanksitem X. Qian is with the Key Laboratory for Intelligent Networks and Network Security, Ministry of Education, Xi’an Jiaotong University, Xi’an 710049, China, also with the SMILES Laboratory, Xi’an Jiaotong University, Xi’an 710049,China, and also with Zhibian Technology Co. Ltd., Taizhou 317000, China (e-mail: qianxm@mail.xjtu.edu.cn, co-corresponding author).\protect% <-this % stops an unwanted space
\IEEEcompsocthanksitem L. Zhu is with the School of Software, Xi’an Jiaotong University, Xi’an 710049, China (e-mail: zhuli@xjtu.edu.cn, corresponding author).\protect
\IEEEcompsocthanksitem Yi Yang is with the Centre for Artificial Intelligence, University of Technology Sydney, Ultimo, NSW 2007, Australia (e-mail: yee.i.yang@gmail.com).\protect}
\thanks{Manuscript received ...; revised ....}}



\markboth{IEEE Transactions on image processing}%
{Shell \MakeLowercase{\textit{Wang et al.}}:Sketch-Guided Scenery Image Outpainting}



\maketitle

\iffalse
\pagestyle{plain}
\onecolumn{
%\renewcommand\onecolumn[1][]{#1}
\maketitle

\begin{center}
%\setlength{\abovecaptionskip}{8pt} 
%\setlength{\belowcaptionskip}{0pt} 
\centering
%\begin{figure*}
%\includegraphics[height=2.4in, width=6.5in]{big_first.png}
\includegraphics[height=3.4in, width=6.8in]{first_page_new.png}
\captionof{figure}{Illustration of the sketch-guided scenery image outpainting. Our proposed method can synthesize the desired outpainting results according to the sketches manually drawn by users.}
%\end{figure*}
\label{first page exhibition}
\end{center}
}
\fi

\begin{abstract}

The outpainting results produced by existing approaches are often too random to meet users’ requirements. In this work, we take the image outpainting one step forward by allowing users to harvest personal custom outpainting results using sketches  as  the  guidance.  To  this  end,  we  propose  an encoder-decoder based network to conduct sketch-guided outpainting, where two alignment modules are adopted to impose the generated content to be realistic and consistent with the provided sketches. First, we apply a holistic alignment module to make the synthesized part be similar to the real one from the global view. Second, we reversely produce the sketches from the synthesized part and encourage them be consistent with the ground-truth ones using a sketch alignment module. In this way, the learned generator will be imposed to pay more attention to fine details and be sensitive to the guiding sketches. To our knowledge,  this  work  is  the first attempt to explore the challenging yet meaningful conditional scenery image outpainting.  We  conduct  extensive  experiments  on  two collected  benchmarks  to  qualitatively  and  quantitatively  validate  the effectiveness  of  our  approach  compared  with  the  other  state-of-the-art generative models.

\end{abstract}


% \fi
\IEEEpeerreviewmaketitle

\section{Introduction}
Image outpainting, also known as image extrapolation, aims at predicting unknown regions beyond the boundary according to currently seen image pixels. Many disparate disciplines demand a strong need for high-quality image extensions. For example, in virtual reality, it is often necessary to simulate different camera extrinsics according to current visual content, which requires making a reasonable extension for the original image.

For an input image, traditional outpainting methods usually focus on designing searching strategies to find regions in a candidate pool ~\cite{T_inpaint1,T_inpaint2,T_inpaint3,T_inpaint4,cmm}. As a consequence, their performances heavily depend on the size of the pool and suffer from limited searching space. Inspired by the success of Generative Adversarial Networks (GANs)~\cite{GAN}, researchers recently propose to synthesize additional contents for the inputs~\cite{boundless,yzx,sabini2018}. However, traditional searching approaches and GAN-based methods both only focus on the authenticity and the semantic consistency between the new content and the original input, and the synthesized random outpainting results are usually below users' expectations. Furthermore, since there are no random variables or control information introduced, existing outpainting systems can only produce one outpainting result for an input image. These limitations make the current outpainting methods unable to meet practical situations. To address the weaknesses of existing outpainting methods, one promising solution is to enable users to acquire personal custom outpainting results, by simply providing guided sketches based on their preferences~\cite{WL1,WL2}, as shown in Fig.~\ref{first page exhibition}. To construct such an outpainting system, several challenges need to be tackled
\begin{itemize}
    \item \emph{Reasonable pixel filling with spatial consistency.} The sketch is a simple and crude clue, which only supplies the desired shape while the corresponding color needs to be adaptively filled by considering the spatial structure information.
    \item \emph{Reasonable synthesis with sketch consistency.} An expected controllable outpainting system is that the synthesized image should exactly match the guiding sketches. Therefore, the learned generator should be sensitive to the users' input and impose the synthesized part to be consistent with the given sketches.
\end{itemize}
\iffalse
\begin{figure*}[t]
%\label{fig2}
%\setlength{\abovecaptionskip}{0pt} 
%\setlength{\belowcaptionskip}{0pt} 
\begin{center}
\subfigure[Free-form inputs]{
    \begin{minipage}[t]{0.315\linewidth}
        \centering
        \includegraphics[width=1.3in]{visual_image/1_1_final1.jpg}\\
        \includegraphics[width=1.3in]{visual_image/1_5.jpg}\\
        \label{fig1:a}
    \end{minipage}
    }
\subfigure[Outpainting results]{
    \begin{minipage}[t]{0.315\linewidth}
        \centering
        \includegraphics[width=1.3in]{visual_image/m0_160.jpg}\\
        \includegraphics[width=1.3in]{visual_image/1_1.jpg}\\
        \label{fig1:b}
        %\caption{fig1}
    \end{minipage}
    }
\subfigure[Original images]{
    \begin{minipage}[t]{0.315\linewidth}
        \centering
        \includegraphics[width=1.3in]{visual_image/1_3_final1.jpg}\\
        \includegraphics[width=1.3in]{visual_image/1_4.jpg}\\
        \label{fig1:c}
        %\caption{fig1}
    \end{minipage}
    }
\caption{Two examples of sketch-guided image outpainting. The first column shows the raw inputs, and the corresponding results are exhibited in the second column, the last column exhibits the original images}
%\vspace{-0.4cm}
\label{fig2}
\end{center}
\end{figure*}
\fi

\begin{figure*}
\centering
%\begin{figure*}
%\includegraphics[height=2.4in, width=6.5in]{big_first.png}
\includegraphics[height=3.4in, width=6.8in]{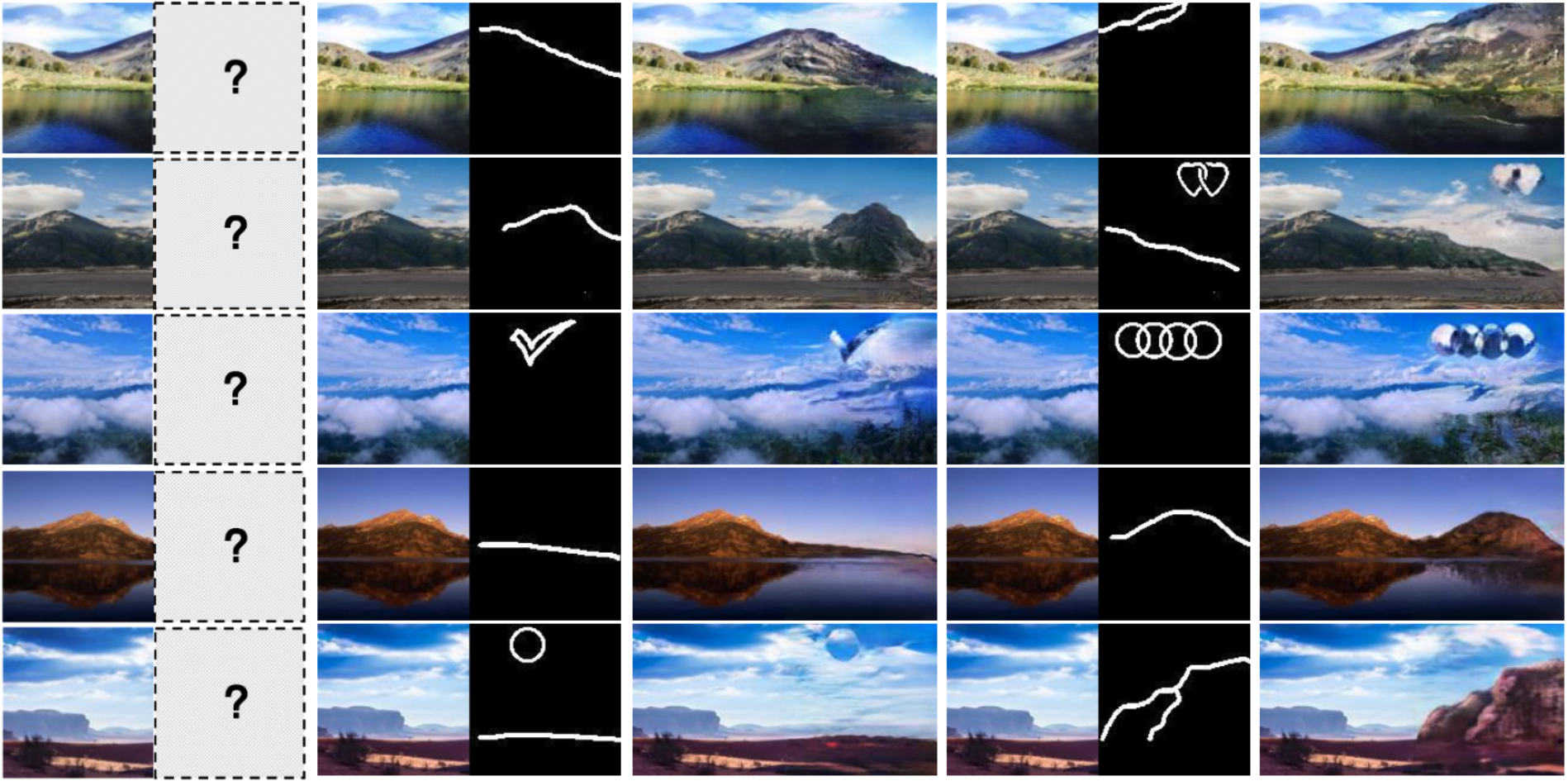}
\caption{Illustration of the sketch-guided scenery image outpainting. Our proposed method can synthesize the desired outpainting results according to the sketches manually drawn by users.}
%\end{figure*}
\label{first page exhibition}
\end{figure*}

To address the above raised challenges and facilitate the Sketch-Guided Scenery Image Outpainting (SGSIO), we contribute a robust system with three basic modules, \ie generator module, holistic alignment module and sketch alignment module. To be specific, we introduce two position channels to the generative module as assistant inputs, which can help the generator perform reasonable pixels filling, by building an explicit link between semantic regions and specific positions. In addition, to guarantee the predicted content exactly matches the users' expectations, we further adopt a conditional skip connection mechanism to prevent the desired sketch from distorting, which is achieved by emphasizing the desired shape in decoding step. To further make the synthesized content be realistic and consistent with the provided sketches, we augment the system with two alignment modules. In general, the holistic alignment module focuses on synthesizing authentic images from a global perspective, while the sketch alignment module attempts to reconstruct the fine details for further enhancing the authenticity. First, the holistic alignment module is employed to generate authentic images by adversarial training, using a global discriminator and a local discriminator for collaboration. In this way, the synthesized part is encouraged to be similar to the real one from its global appearance.
Second, we introduce an additional sketch alignment module to help the network rebuild image details by enforcing the generator to reconstruct the high-frequency information in images, imposing the generator to pay more attention to the detailed sketches and produce reasonable new content with sketch consistency accordingly. \yx{Several examples generated by the proposed outpainting system are shown in Fig.~\ref{first page exhibition}, where the first column is the input image for outpainting. The second to fifth columns subsequently show the two types of guiding sketches and the corresponding outpainting results, and every two columns form a pair of results, \ie, the inputs and the outputs.} For the input images, our system allows users to edit the predicted content by free-style sketches based on their preferences. Users may expect special clouds with specific shape  or control the mountain trend, with our proposed system, these expectations can be easily achieved by feeding the manually drawn sketches to guide the synthesis. It can be observed that our system can smoothly produce the synthesized part regarding the consistency from both the given sketches and the surrounding contextual information.

%When the training finished, our system can synthesize the desired image according to the input and the guiding sketch. As shown in the bottom row of Fig.~\ref{fig2}, the user inputs an image that shows a river, and expects the river to flow through the generated image instead of being blocked by land or mountain. With the help of our designed network, the expectation can be achieved by simply providing a sketch to guide the generation just as shown in Fig.~\ref{fig2}.

This work makes the first attempt to conduct conditional scenery image outpainting. Our proposed method not only makes it possible for users to control the image extrapolation but achieves state-of-the-art performances on NS6K dataset~\cite{yzx}. To evaluate the robustness and test the performance under more complex scenarios, we further build a new dataset called NS8K by removing some similar images from NS6K and absorb additional thousands of images with diverse appearances from world-wide famous scenic spots. Our method is more outstanding on both two datasets. The
contributions of this paper can be summarized as follows:
\begin{itemize}
\item We consider a new outpainting task that allows users to control the scenery image outpainting by free-form sketches, which is an under explored task. We hope this work can serve as a solid baseline and ease future research for conditional image outpainting.

\item We develop an outpainting system with one module for generation and two modules for appearance alignment. With the assistant of the proposed modules, our outpainting system can pay more attention to fine details and successfully produce sketch-consistent outputs.
%A conditional skip connection is designed to make the synthesized part spatial consistent with the guiding sketch. What's more, a random mask mechanism and a seaming loss are proposed to alleviate the overfitting and smooth the transition boundary.
%, which can help synthesize more smooth and realistic results. 

\item  We contribute a natural image dataset NS8K. This new dataset contains much more complex and diverse images than the original NS6K. 
\end{itemize}

The reminder of this paper is organized as follows: In section~\ref{related_work}, we review the related works of this paper. Section~\ref{method} elaborates the details of our proposed sketch-guided image outpainting network. The expeirments are shown in section~\ref{experiment}. Finally, conclusion and future works are given in section~\ref{conclusion}.

\begin{figure*}
\centering
%\begin{figure*}
%\includegraphics[height=2.4in, width=6.5in]{big_first.png}
\includegraphics[height=2.2in, width=6.2in]{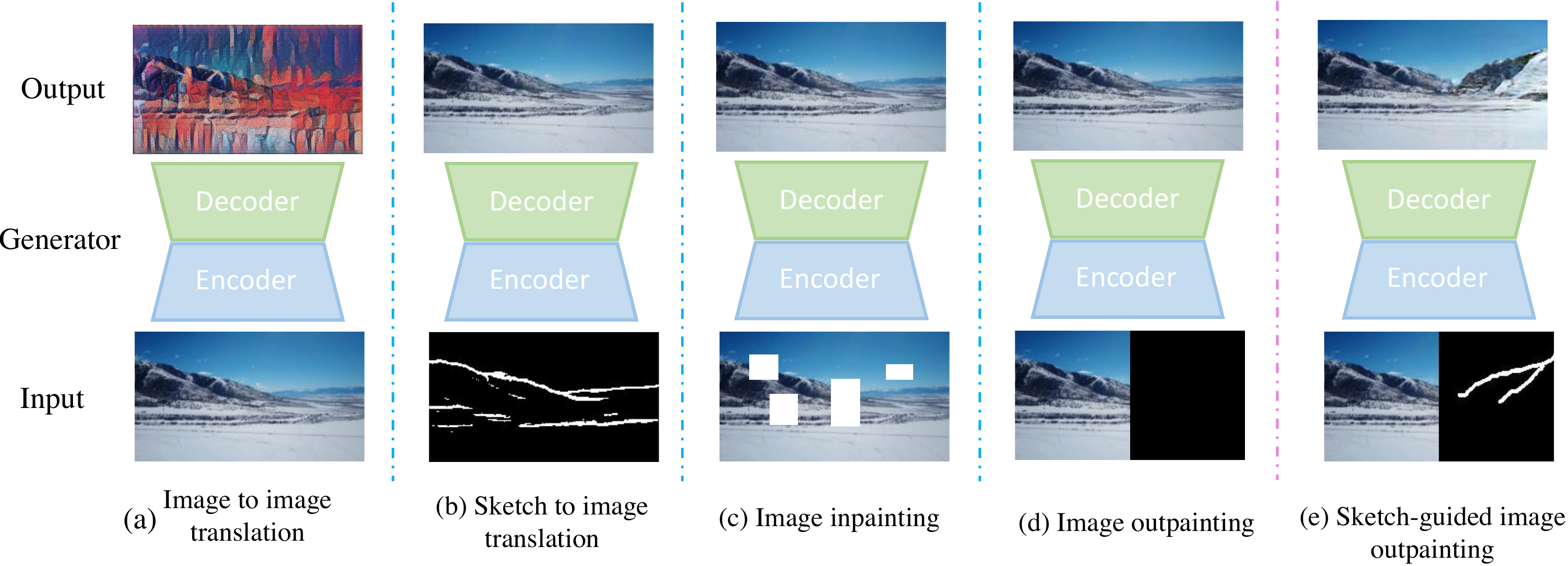}
\caption{A toy comparison of the difference and the relation between sketch-guided image outpainting and existing tasks including image to image translation, sketch to image translation, image inpainting, and image outpainting.}
%\end{figure*}
\label{diff_rela}
\end{figure*}

\section{Related work}
\label{related_work}
%\textcolor{red}{The sketch-guided image outpainting attracts few attention these years, however, there are still many problems that are related to this challenge yet meaningful task. In this section, we would subsequently review the existing works in three sub-field: image-to-image translation, image inpainting and image outpainting.}
The sketch-guided image outpainting attracts few attention these years. In this section, we would subsequently review the existing works in three sub-field: image-to-image translation, image inpainting and image outpainting. \yx{Fig.~\ref{diff_rela} gives a toy example to illustrate the differences of existing works and the sketch-guided image outpainting.} 
\subsection{Image to Image Translation} 
The image to image (I2I) translation attempts to map the images in one domain to another~\cite{I2I_2017, DCGAN, hu}, since the I2I problem was proposed by Isola \etal~\cite{I2I_2017}, it has attracted much attention due to the well generality for many downstream tasks. Isola \etal make the first attempt and design a general solution, \ie Pix2Pix, for tacking image-to-image translation~\cite{I2I_2017}. Zhu \etal cyclically synthesize the source map and target image, they propose a novel network named CycleGan to improve the quality of  results~\cite{CycleGan}.  The classic structures, Pix2Pix and CycleGan, design simple structures but result in very promising results, which motivate more researchers to employ the I2I framework as their basic architectures. To be specific, Yu \etal employ the I2I architecture and design a multi-mapping style transfer framework~\cite{multiMP2019}. Luan \etal capture the style from a reference image and transfer the style to the target one~\cite{Luan_2017_CVPR}. In~\cite{sketchGan} and~\cite{sketchGan1}, the authors model the sketch to image synthesize as an image to image translation framework, the developed network could synthesize the realistic images from the input sketches. Lu \etal and Madam \etal employ the I2I framework and restore the sharp images from the corresponding blurred ones~\cite{boyu_deblurring, deblur1}. 
Besides the style transfer, sketch to image synthesise and image debulrring, many tasks like image inpainting~\cite{Yu0,Yu_2018_CVPR}, image denoising~\cite{denoise1, denoise2,imRes,Noise}, and image super-Resolution~\cite{super1,super2,super} all attempt to design solutions based on I2I framework. However, I2I methods only focus on synthesizing authentic images while pay no attention to the semantic and stylistic consistency between the input and the generated part. Consequently, it will acquire a poor performance if we stiffly extend the I2I methods to conduct image outpainting~\cite{yzx,boundless}. 

%The classic I2I works like Pix2Pix\cite{I2I_2017} and CycleGAN~\cite{CycleGan} design simple structure but produce very promising results. Many problems employ the I2I framework as their fundamental architecture, such as image style transfer~\cite{multiMP2019,Luan_2017_CVPR},  sketch to image synthesis~\cite{sketchGan,sketchGan1}, image inpainting~\cite{Yu0,Yu_2018_CVPR} and even image deblurring~\cite{boyu_deblurring,deblur1}. However, I2I methods only focus on synthesizing authentic images while pay no attention to the semantic and stylistic consistency between the input and the generated part. Consequently, it will get a poor performance if we stiffly extend the I2I methods to conduct image outpainting~\cite{yzx,boundless}.

%on the subject of image outpainting, it will get a poor performance if we stiffly extend the I2I methods to conduct image outpainting~\cite{yzx, boundless}.  This is because one of the main obstacles in outpainting task is the semantic and style consistency between the input and the generated part, while the traditional I2I methods pay no attention to it and the generated images are semantically different with the original input~\cite{boundless,yzx}.  
\begin{figure*}[t]
    \centering
    \includegraphics[height=2.7in, width=6.8in]{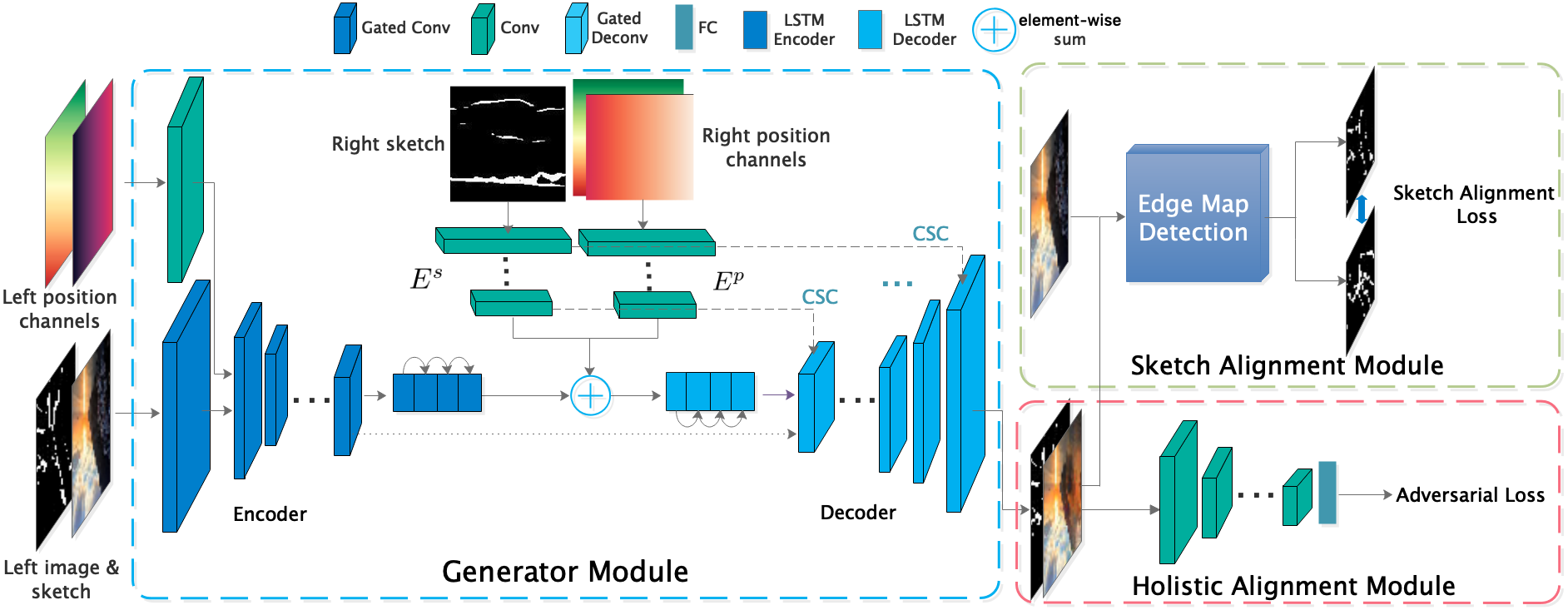}
    \caption{The architecture of the proposed method. Our framework consists of three modules. The generator module takes the image and sketch as inputs and predicts new content beyond the boundary for the input image. The holistic alignment module is responsible to discriminate whether the synthesized part is fake or not from a global view, and the sketch alignment module focuses on imposing the generator to be sensitive to the fine details and recovering the high-frequency information to boost the outpainting quality.}
    \label{fig:workflow}
\vspace{2pt}
\end{figure*}

\subsection{Image Inpainting} Image inpainting targets at reconstructing missing areas in the corrupted images ~\cite{guozongyu0, Xie0, Pathak_2016_CVPR, Yang_2017_CVPR}, which has been well explored. Extensive efforts have been dedicated to this field. Yu \etal design a contextual attention module and propose to compensate the missing areas using the pixels from similar regions~\cite{Yu_2018_CVPR}. Iizuka \etal employ a fully-convolutional neural network and use global and local context discriminators to train an inpainting system~\cite{GL}. Recently, the focused case for image inpainting has been moving from predicting the missing region with formal shape to the irregular inpainting ~\cite{Yu0,Yu_2018_CVPR,inpaint_xie,guozongyu0,Liu0}. Liu \etal propose a partial convolution, which could progressively predict the missing pixels from the surrounding content~\cite{Liu0}. Yu \etal develop a gate convolution to adaptively learn a soft mask, and the designed architecture could significantly improve the inpainting results. In~\cite{inpaint_xie}, Xie \etal attempt to the learn a feature re-normalization by the desinged learnable attention map module, which could effectively adapt to the irregular holes. Guo \etal propose a full-resolution residual network (FRRN) to fill irregular holes, the authors aim at compensating more textural details for the damaged areas~\cite{guozongyu0}. Han \etal propose a two-stage image-to-image generation framework, which could perform compatible and diverse inpainting~\cite{inpaint_add1}.  In~\cite{inpaint_add2}, Ren \etal split the inpainting task as two part, \ie structure reconstruction and texture generation, and the authors design a two-stage framework to yield texture-detailed results. The classic methodology of image inpainting predicts the damaged pixels from the neighbors based on the convolutional operation. These methods could make a success on image inpainting, however, they suffer from the lack of surrounding pixels when applied on outpainting task ~\cite{boundless,yzx}. Comparing to image inpainting, it is an extra challenge for outpainting that the missing region is relatively large and far away from the valid pixels. 

\iffalse
the focused case has been moving from predicting the missing region with formal shape to irregular inpainting ~\cite{Yu0,Yu_2018_CVPR,inpaint_xie,guozongyu0,Liu0,GL}.  The classic methodology predicts the damaged pixels from the neighbors based on the convolutional operation. 
%Parial convolution ~\cite{Liu0} design a mask covolution to make the output condition only on the valid mask and feature re-normalization is further employed to scale the output. Gated Convolution proposed by Yu et al. ~\cite{Yu0} introduces a dynamic gate structure to adaptively learn a soft mask. Inspired by the partial convolution, Guo et al. ~\cite{guozongyu0} design a Full-Resoultional Residual Block (FRRB) and iteratively prediction the missing pixels by stacking the FRRB module. 
These methods could make a success on image inpainting, however, they suffer from the lack of surrounding pixels when applied on outpainting task ~\cite{boundless,yzx}. Comparing to image inpainting, it is an extra challenge for outpainting that the missing region is relatively large and far away from the valid pixels. 
\fi

\subsection{Image Outpainting} Traditional outpainting methods first search relevant patches from a pre-defined candidate pool, and then the retrieved patches are stitched with the input image to conduct extrapolation~\cite{T_inpaint1,T_inpaint2,T_inpaint3,T_inpaint4}. Zhang \etal formulate the outpainting in the shift-map image synthesis framework, the authors search a guide image and analyze the self-similarity of the guide image to generate the allowable local transformations, which is then applied to the input image to conduct extrapolation~\cite{T_inpaint3}. Wang \etal  use the library images to determine the consistent content for the regions and propose a data-driven approach to extrapolate the image to a given distinctly large one~\cite{T_inpaint4}. The performance of the traditional search-based methods depends on the searching results and the number of candidate images. Awful patch choosing would cause poor performance. These methods are not flexible enough and hard to extend to more complex situations. Inspired by the success of the generative adversarial networks (GANs), recently researchers utilize GANs framework to synthesize new contents beyond the boundary~\cite{boundless,yzx,sabini2018}. Yang \etal utilize GANs and the recurrent neural network to iteratively predict new contents for current region. Teterwak \etal design a powerful discriminator that takes the groundtruth image as input to form an improved inception loss, the developed network could efficiently restore the images. In~\cite{outpaint_add1}, Zhang \etal study a special outpainting task, aiming to generate a set of realistic backgrounds with a given small foreground region. Wang \etal allow the users to control the margin of the boundary, and propose to synthesize the news contents matching the expected resolution~\cite{outpaint_add2}. Wu \etal explore the outpainting problem for the portrait image, the authors design a two-stage framework to produce realistic portrait image~\cite{outpaint_add3}. However, existing outpainting methods only focus on generating realistic images but introduce no extra information to guide the final synthesis. As a consequence, these works all produce random results.

%Sanini et al. ~\cite{sabini2018} train a DCGAN ~\cite{DCGAN} based structure by a three-phase training schedule for image outpainting. Teterwak et al. ~\cite{boundless} introduce a semantic conditioning discriminator to extend the pixel beyond the boundary.  Different from the papers above, Yang et al. ~\cite{yzx} propose a encoder-decoder architecture, the information in left half image is transferred to decode the right half by a RNN module, which has shown promising results.  

\section{Methodology}
\label{method}
\subsection{Overview}
As shown in Fig.~\ref{fig:workflow}, our framework comprises of three modules, \ie the generator module, the holistic alignment module and the sketch alignment module. Our generator module takes an image and its sketch as inputs, and synthesizes additional right half content using its right counterpart as the guidance. The holistic alignment module is responsible to predict a scalar, which coarsely indicates the input is generator-produced or real from a global view. In contrast, the sketch alignment module aims at pursuing the detailed agreement between the sketches inferred from the synthesized part and the ground-truth. These three modules are jointly trained by the classic adversarial loss~\cite{GAN} and the proposed sketch alignment loss. During the training process, our system takes the left half image, the left half sketches, and the right half sketches as inputs to reconstruct the entire image. At the testing stage, users could feed an image and the free-form guiding sketches, to synthesize the desired image with additional right half, as shown in Fig.~\ref{first page exhibition}. In the following, we will first introduce each module one by one, and provide the training details subsequently.

%We give the overview of our proposed approach in Fig.~\ref{fig:workflow}. In general, our proposed network is an encoder-decoder architecture. The encoder compresses the inputs to hidden features, and the decoder takes the encoded features and the corresponding guiding information as inputs to predict the complete image. Meanwhile, the conditional skip connection is applied to make the final image exactly match the guiding sketch. In the training procedure, the seaming loss and the random mask mechanism are further employed to learn a more robust generation model. More details of each component are given in the following.

\subsection{Generator Module}\label{section3-1}
Following the previous state of the art outpainting method~\cite{yzx}, our generator also takes an encoder-decoder based network. The encoder compresses the inputs to hidden features, and a LSTM~\cite{LSTM} encoder collaborating with a LSTM decoder are employed to predict the hidden features of the complete image, which is further fed into the subsequent decoder layers to restore the complete image. 

\begin{figure*}[t]
\begin{center}
     \subfigure[The inputs]{
    \begin{minipage}[t]{0.20\linewidth}
    \includegraphics[width=1.56in]{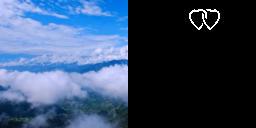}\\
    \vspace{-0.4cm}
    \includegraphics[width=1.56in]{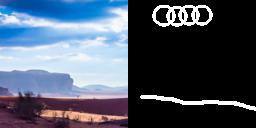}\\
    \vspace{-0.3cm}
    \end{minipage}
    }
   \hspace{0.01cm}
    \subfigure[Results W/O CSC]{
    \begin{minipage}[t]{0.20\linewidth}
    \includegraphics[width=1.56in]{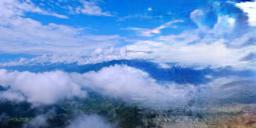}\\
    \vspace{-0.4cm}
    \includegraphics[width=1.56in]{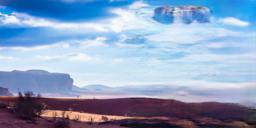}\\
     \vspace{-0.3cm}
    \end{minipage}}
    \hspace{0.12cm}
    \subfigure[Results with CSC]{
    \begin{minipage}[t]{0.20\linewidth}
    \includegraphics[width=1.56in]{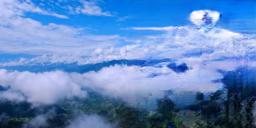}\\
    \vspace{-0.4cm}
    \includegraphics[width=1.56in]{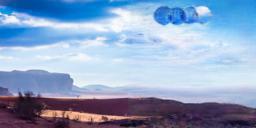}\\
     \vspace{-0.3cm}
    \end{minipage}
    }
    \hspace{0.01cm}
    \subfigure[The original images]{
    \begin{minipage}[t]{0.20\linewidth}
    \includegraphics[width=1.56in]{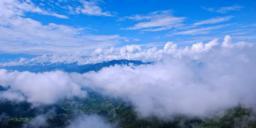}\\
    \vspace{-0.4cm}
    \includegraphics[width=1.56in]{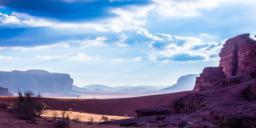}\\
     \vspace{-0.3cm}
    \end{minipage}
    }
\end{center}
\vspace{-0.3cm}
 \caption{The generator without CSC can not ensure the results exactly meet the guiding sketches. When the guiding sketch is small, the model may totally `forget' the sketch in decoding step.}
 %\vspace{-0.2cm}
\label{csc_abalation}
\end{figure*}

Compared to synthesize random contents, sketch-guided outpainting poses two extra challenges for the generator. First, with limited training samples, the learned outpainting model is often hard to cover all the free-style sketches drawn by users, making it not robust to those novel sketches. Since it is not practical to augment the training with a huge number of images with various sketches, one promising solution is to impose the filled pixels around the given sketches to be consistent with the contextual information and the learned prior knowledge. Intuitively, \emph{cloud} and \emph{sky} usually appear at the upper part of one scenery image while \emph{grass} is the opposite. If the learned outpainting model could be well equipped with position-aware prior knowledge, it will take \emph{white} or \emph{blue} colors to fill those novel sketches given at the upper part rather than the \emph{green} one. Motivated by the relation cues between the pixels and the position, we introduce two position channels, to help the generator learn position-aware knowledge and be robust to predict the filled pixels for novel sketches. Second, we should ensure the synthesized image exactly match the guided sketches, since we find the generator often produce deformed results comparing to the given sketches in our experiments. For some extreme cases, the generated content even totally loses the shape information for relative small guiding sketches, as shown in Fig.~\ref{csc_abalation} (b). 
Consequently, the synthesized results do not well match users' intentions. To address this issue, we design a conditional skip connection to emphasize the desired shape in the decoding stage, which can effectively help the generator ``remember" the information of the expected shape. 

% Compared to synthesize random contents, sketch-guided outpainting poses two extra challenges for the generator. Filling the reasonable pixels for the drawn sketches poses the first challenge. The difficulties stem from the free-style sketches may be much different from the training data, therefore, the generator is hard to generalize to diverse free style sketches.  Motivated by the relation cues between the pixels and the position, we introduce two position channels help our generator to predict reasonable pixels. 

% The second challenge is to ensure the synthesized image exactly match the guiding sketch, since we find the generator may produce deformed results comparing to the guiding sketch in our experiment, and even totally loses the shape information for relative small guiding sketch, as shown in Fig.~\ref{csc_abalation} and Fig.~\ref{ablation_free}. Consequently, the synthesized results do not meet users expection. To address this issue, we design a conditional skip connection to emphasize the desired shape in the decoding stage, which attempts to help the generator "remember" the information of the expected shape.  

\subsubsection{Position Channels}
%Yu et al.~\cite{Yu0} directly feed image and sketch to the generator to train their inpainting system. However, when it encounters sketch-guided outpainting, the situation becomes challenging because of the diverse free style sketches and the lack of the contextual information from surrounding pixels. For instance, when a user draws some lines as shown in Fig.~\ref{fig2}, there is no available neighbor to help predict the pixels of different sides. 
Intuitively, for a scenery image, different types of objects should appear in specific positions, \eg the \emph{clouds} should locate in the sky (top part) instead of the ground (bottom part), while the \emph{lake/land/rock} are more likely to be in the bottom part. Therefore, for scenery image outpainting, the position relation between the semantic region and the specific position is a helpful clue for new content prediction, what's more, the learned positional prior knowledge would play a key role in helping the network robustly generalize to the free-style outpainting. Inspired by the above considerations and the successes of position maps on position-sensitive tasks~\cite{posiConv}, we utilize two additional position channels, \ie the width channel $X\in \mathbb{R}^{H\times W\times 1}$ and the height channel $Y\in \mathbb{R}^{H\times W\times 1}$, to assist our system in predicting reasonable pixels for the outpainting:
\begin{equation}\label{eq.2}
\left\{ \begin{aligned}
    X(i, j) = \dfrac{2\times j - W}{H}, &  & 0\leq j \leq W-1,\\
    Y(i, j)  =\dfrac{2\times i - H}{H}, &  & 0\leq i \leq H-1,
\end{aligned}
\right.
\end{equation}
where $X(i,j)\in [-W/H, (W-2)/H]$ and $Y(i,j)\in [-1,(H-2)/H]$, $W, H$ are the width and height of image in training set, respectively. The values of two position channels range in different intervals while maintain the same changing step, \ie $1/H$. 

The convolutional neural network can capture some position cues by enlarging receptive field~\cite{vgg16}, however, the captured position information is implicit and not powerful enough to benefit the overall outpainting. Different from the implicit position cues from CNN, our position channels attempt to model the explicit relation between the pixel semantics and its position, and provide more forceful assistant information, to help the generator predict reasonable pixels, especially for the free-form outpainting.
%Let $I^l\in \mathbb{R}^{H\times W/2\times 3}$ and $S^l\in \mathbb{R}^{H\times W/2\times 1}$ be the left half image and sketch, 

%The CNN can capture position cues by enlarging receptive field~\cite{vgg16}, however, the captured position cues are implicit and not powerful enough, by introducing the position channels (PCs), the relation between pixel semantics and its position can be explicitly linked. 

%\begin{equation}\label{eq.3}
%    f = E(I^l,S^l, X^l, Y^l; \theta_e),
%\end{equation}
%where the $X^l$ and $Y^l$ are the corresponding left half of width and height channels respectively, $f\in  \mathbb{R}^{M\times K\times C}$ are the encoded hidden features, and $\theta_e$ is the learnable parameters in encoder $E$ . 

In practice, the position channels are also split into two parts along with the width dimension, \ie the left part $X^l, Y^l \in \mathbb{R}^{H\times W/2\times 1}$ and the right part $X^r, Y^r \in \mathbb{R}^{H\times W/2\times 1}$, to fit the generator architecture. The left two position channels are first encoded by a convolution layer and then concatenated with the fused features from the input image and sketch, which is further encoded to obtain the final left hidden representations. In the following decoding step, the right half position channels and the right half sketch are first encoded by a position encoder $E^p$ and a sketch encoder $E^s$, respectively. Then, the compressed two types of representations are element-wise added with the sequential features from the LSTM encoder, to serve as the initial state for the following LSTM decoder whose responsibility is to predict the hidden features of the full image. The subsequent decoder module takes the full hidden features to rebuild the complete image, by a series of convolution and upsampling operations.

\subsubsection{Conditional Skip Connection} To synthesize the expected image that exactly matches the guiding sketch, we design a conditional skip connection structure inspired by U-net~\cite{u-net} to emphasize the desired shape in each decoding step. Let $s^i\in \mathbb{R}^{h_i\times w_i\times c^s_i}$ and $p^i\in \mathbb{R}^{h_i\times w_i\times c^p_i}$ be the outputs of the $i$-th layer in $E^s$ and $E^p$, respectively, the right half features of output in the $i$-th decoding layer are denoted as $d^i\in \mathbb{R}^{h_i\times w_i\times c^d_i}$. These three tensors are first channel-wise concatenated and then fed forward through three convolution layers with kernel size $1\times 1$, $3\times 3$ and $1\times 1$, to get new features with the same shape as $d^i$. To make the training more stable, we also introduce
a residual connection to conduct an element-wise addition between the new features and $d^i$ to get the final output of the CSC module $\check{d}^i\in \mathbb{R}^{h_i\times w_i\times c^d_i}$. And the output of the $i$-th decoding layer is updated by replacing the original right half feature $d^i$ with $\check{d}^i$ accordingly. 

The difference between the CSC and the U-net is two-fold: First, the CSC only focuses on the right half of the decoder feature which corresponds to the guiding sketch and the corresponding position region. Second, two components of connection in CSC are not symmetrical, features from the condition encoding modules contain the guiding information and the assistant position features,  while features in decoding layers encode the additional visual feature transferred from the inputs. In our experiment, the CSC can not only improve the free-form oupainting, but speed up the network convergence as shown in Fig.~\ref{converge}.

\subsection{Holistic Alignment Module}
%Following the same strategy in~\cite{Yu_2018_CVPR} and~\cite{yzx}, our discriminator consists of two modules. The local discriminator discriminates whether the synthesized part is generator-produced or real, while the global one determines whether the entire image is real or not. Both of them take image and sketch as inputs and output a 1-D scalar by several striding convolution layers and a fully connected layer.

The holistic alignment module, which is responsible for discriminating the input image is fake or real, is introduced to conduct adversarial learning. Following the same strategy in~\cite{Yu_2018_CVPR} and~\cite{yzx}, our holistic alignment module consists of two discriminators. The local discriminator discriminates whether the synthesized part is generator-produced or real, while the global one determines whether the entire image is real or not. Both of them take the concatenation of image and sketch as inputs and output a 1-D scalar by several striding convolution and a fully connected layer.

The overall architecture employs the Wasserstein GANs~\cite{wgan} framework, and the network parameters are trained by solving the min-max optimization:
%\max_{\textit{\textbf{G}}}
\begin{equation}
\label{d_loss}
\begin{aligned}
    \mathcal{L}_d = &\min_{\textit{\textbf{D}}}\underset{\hat{I}\sim \mathbb{P}_f}{\mathbb{E}}[\textit{\textbf{D}}(\hat{I}, S)]-\underset{I\sim \mathbb{P}_r}{\mathbb{E}}[\textit{\textbf{D}}(I, S)]\\
    &+ \lambda_w\underset{\tilde{I}\sim \mathbb{\tilde{P}}}{\mathbb{E}}[(||\nabla_{\tilde{I}}\textit{\textbf{D}}(\tilde{I}, S)||_2-1)^2],\\
\end{aligned}
\end{equation}

\begin{equation}
  \hspace{-0.62in}   \mathcal{L}_g = \min_{\textbf{\textit{G}}}-\underset{\hat{I}\sim \mathbb{P}_f}{\mathbb{E}}[\textbf{\textit{D}}(\hat{I}, S)].
\end{equation}
where $\textit{\textbf{D}}$ is the  discriminator and $\textit{\textbf{G}}$ indicates the generator,  $\mathcal{L}_d$ and $\mathcal{L}_g$ are the discriminator loss and generator loss, respectively. The last term in Eq.~\ref{d_loss} is the gradient penalty to enforce the Lipschitz constraint ~\cite{wgan}, $\tilde{I}$ is a random sample from a probability distribution $\tilde{\mathbb{P}}$. 

Besides the adversarial loss, the reconstruction loss is also implemented as a masked $\ell_1$ loss, which optimizes for coarse image agreement:
\begin{equation}
    \mathcal{L}_1 = M\odot||I - \hat{I}||_1,
\end{equation}
where $M$ is a mask proposed by Yang et al.~\cite{yzx} to reduce the weight of reconstruction loss along the prediction direction, $\odot$ indicates the element-wise product and $I$ is the groundtruth image.

\subsection{Sketch Alignment Module}
The holistic alignment module only focuses on the overall image and the synthesized part, and pays less attention to the details of the synthesized content. Consequently, the outpainting model trained with the holistic alignment module could successfully restore most of the low-frequency information but fail to well keep the high-frequency details, leading to the blurry boundary between different semantic regions. 

To further enhance the outpainting quality, we augment the system with a sketch alignment module to restore the high-frequency information. To be specific, with the generated synthesized part, we first leverage an %unsupervised 
edge detector~\cite{HED} to reversely produce its edge map where high-frequency information is maintained. Then, the sketches from input are adopted as ground-truth and a sketch-based alignment loss is applied to impose the inferred edge map to be consistent with the ground-truth one. Formally, let $\hat{S}$ be the sketch of the synthesized image from generator, which can be obtained by feeding the image rebuilding $\hat{I}$ through the edge detector, and $S$ is the sketch map from ground-truth, our sketch-based alignment loss is defined as follow:

%To further enhance the outpainting quality, we propose a sketch alignment loss to restore the high-frequency information,  based on the image edge map because of its effectiveness in capturing the high frequency information of images. Let $\hat{S}$ be the sketch of the synthesized image from generator, which can be obtained by feeding the image rebuilding $\hat{I}$ through the edge detection module, and $S$ is the sketch of the groundtruth image, our sketch alignment loss is defined as follow:
\begin{equation}
\label{sk_alignment}
    \mathcal{L}_{s} = ||\hat{S} - S||^2_2.
\end{equation}
where $||\cdot||_2$ indicates the $l_2$ loss.  The reconstruction defined by Eq.~\ref{sk_alignment} enforces the generator to recover the high frequency details in the groudtruth image, it is also in line with our sketch-guided setting.

\subsection{Training}\label{section3-3}

%Besides the commonly used reconstruction loss and the adversarial loss, we further design a seaming loss to smooth the transition between the original half and the synthesized part. In our architecture, the output $O^l_K$ from RNN encoder $R^e$ encodes the feature around the boundary from left to right direction. An intuition is that the feature of the generated image close to the boundary should be similar to the left one, which could contribute to the smooth transition of the connection region.  Motivated by this, we reverse the right half of $\hat{I}$, the corresponding sketch and position channels. And following the same operations in processing the left half inputs, the features near the boundary along right to left direction can be obtained as follow:
%\begin{equation}
%    o^{ir}_k,h^{ir}_k = R^e(f^{ir}_{k-1}, h^{ir}_{k-1};\theta^R_e), \quad 1\leq k\leq K
%\end{equation}
%where $f^{ir}=[f^{ir}_0, f^{ir}_1, \dots, f^{ir}_{K-1}]$ are the encoded features from the encoder module $E$ by feeding the inverted synthesized right half image, the right sketch and corresponding position channels, the initial hidden state $h^{ir}_{0}$  is set as zero tensor. Thus, our seaming loss is defined as:
%\begin{equation}
%    \mathcal{L}_{s} = ||o^l_K - o^{ir}_K||^2_2.
%\end{equation}

\setlength{\tabcolsep}{1mm}{
\begin{table}[t]
\setlength{\abovecaptionskip}{0pt}%    
\setlength{\belowcaptionskip}{2pt}%
\begin{center}
\begin{tabular}{l|c|c|c}
\hline
Layer & Input & Output & Kernel size \& Strides\\
\hline
 G-Conv &128x128x4 & 64x64x64 & 4x4, strides=2\\
 Conv & 128x128x2 & 64x64x64 & 4x4, strides=2\\
\hline
 G-Conv & 64x64x128 & 32x32x128 & 3x3, strides=2\\
\hline
 G-Conv & 32x32x128 &  16x16x256 & 1x1, strides=2\\
\hline
 G-Resblockx3 & 16x16x256 & 16x16x256 & 1x1,3x3,1x1, strides=1\\
\hline
 G-Conv & 16x16x256 & 8x8x512 & 3x3, strides=2\\
\hline
 G-Resblockx4 &8x8x512 &8x8x512 & 1x1,3x3,1x1, strides=1\\
\hline
 G-Conv & 8x8x512 & 4x4x1024 & 3x3, strides=1\\
\hline
 G-Resblockx5 & 4x4x1024 & 4x4x1024 &1x1,3x3,1x1, strides=1\\
\hline
 Conv & 4x4x1024 & 4x4x512 & 3x3, strides=1\\
\hline
 LSTM Encoder & 4x(4x512) & 1x(4x512) & -\\
\iffalse
10-2 Sketch Encoder & 128x128x1 & 64x64x64 & 4x4, stirdes=2\\
  &64x64x64 & 32x32x128 & 4x4, strides=2\\
  &32x32x128 & 16x16x256 & 4x4, strides=2\\
  &16x16x256 & 8x8x512 & 4x4, strides=2\\
  &8x8x512 & 4x4x512 &4x4, strides=2\\
  &4x4x512 & 4x2x512 &4x4, strides=(1,2)\\
  &2x4x512 & 4x1x512 & 4x4, strides=(1,2)\\
\fi
 Sketch Encoder & 128x128x1 & 1x(4x512) & -\\
 Position Encoder & 128x128x2 & 1x(4x512) & -\\
\hline
 Sum & 1x(4x512) x 3&1x(4x512) & None\\
\hline
 LSTM Decoder & 1x(4x512) & 4x4x512 & -\\
\hline
 Concat & (4x4x512)x2 & 4x8x512 & None\\
\hline
 G-Resblockx2& 4x8x512 & 4x8x512 & 1x1,3x3,1x1, strides=1\\
\hline
 CSC+G-DeConv & 4x8x512 & 8x16x512 & 3x3\\
\hline 
 G-Resblockx3 & 8x16x512 & 8x16x512 & 1x1,3x3,1x1, strides=1\\
\hline
 CSC+G-DeConv & 8x16x512 & 16x32x256 & 3x3\\
\hline
 G-Resblockx4 & 16x32x256 & 16x32x256 & 1x1,3x3,1x1, strides=1\\
\hline
 CSC+G-DeConv & 16x32x256 & 32x64x128 & 3x3\\
\hline 
 G-DeConv & 32x64x128 & 64x128x64 & 3x3\\
\hline
 G-DeConv & 64x128x64 & 128x256x3 & 3x3\\
\hline
\end{tabular}
\end{center}
%\vspace{-0.2cm}
\caption{The architecture of our generator, where the G-Conv and G-DeConv indicate the gated convolution and the gated deconvolution~\cite{Yu0}, respectively.  G-Resblock refers to the resblock~\cite{resnet} whose convolution operations are replaced by the gated convolutions.}
\label{architecture}
\vspace{-0.3cm}
\end{table}
}

\begin{figure}[t]
    \centering
    \includegraphics[height=2.0in, width=3.5in]{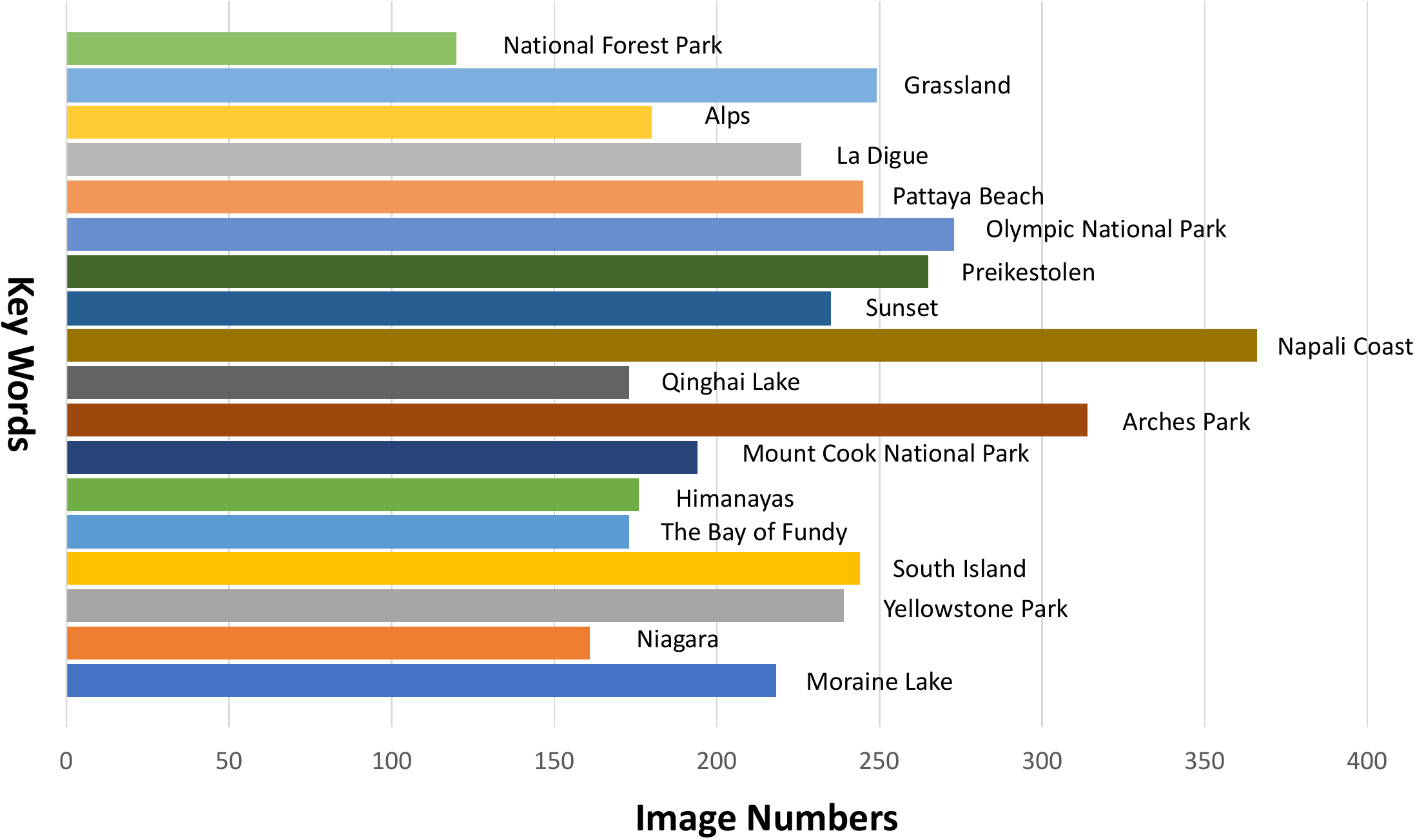}
    %\vspace{-0.2cm}
    \caption{The image distribution for each keyword in collected samples of NS8K dataset, where x-axis indicates the numbers and y-axis shows the keywords.}
    %\vspace{-0.6cm}
    \label{sample_dis}
\end{figure}

\begin{figure*}[t]
\begin{center}
    \subfigure[Pix2Pix~\cite{I2I_2017}]{
    \begin{minipage}[t]{0.18\linewidth}
    \includegraphics[width=1.35in]{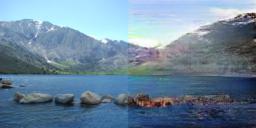}\\
    \vspace{-0.4cm}
    \includegraphics[width=1.35in]{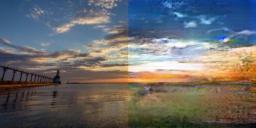}\\
     \vspace{-0.4cm}
    \includegraphics[width=1.35in]{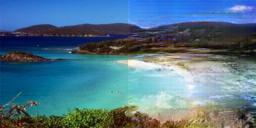}\\
    \vspace{-0.3cm}
    \end{minipage}
    }
   \hspace{-0.18cm}
    \subfigure[NSIO~\cite{yzx}]{
    \begin{minipage}[t]{0.18\linewidth}
    \includegraphics[width=1.35in]{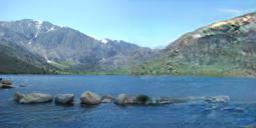}\\
     \vspace{-0.4cm}
    \includegraphics[width=1.35in]{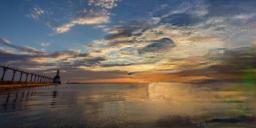}\\
     \vspace{-0.4cm}
    \includegraphics[width=1.35in]{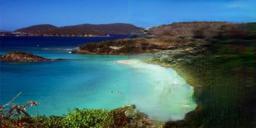}\\
     \vspace{-0.3cm}
    \end{minipage}}
    \hspace{-0.06cm}
    \subfigure[BDIE~\cite{boundless}]{
    \begin{minipage}[t]{0.18\linewidth}
    \includegraphics[width=1.35in]{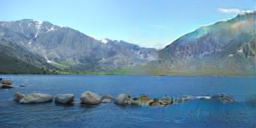}\\
     \vspace{-0.4cm}
    \includegraphics[width=1.35in]{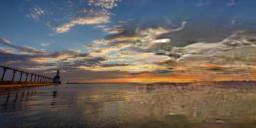}\\
     \vspace{-0.4cm}
    \includegraphics[width=1.35in]{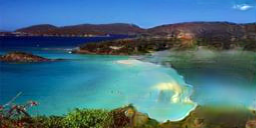}\\
     \vspace{-0.3cm}
    \end{minipage}
    }
    \hspace{-0.18cm}
    \subfigure[Ours]{
    \begin{minipage}[t]{0.181\linewidth}
    \includegraphics[width=1.35in]{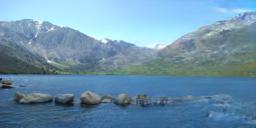}\\
     \vspace{-0.4cm}
    \includegraphics[width=1.35in]{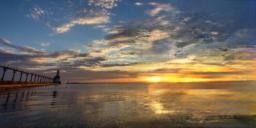}\\
     \vspace{-0.4cm}
    \includegraphics[width=1.35in]{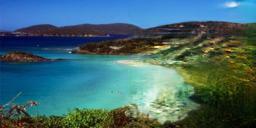}\\
     \vspace{-0.3cm}
    \end{minipage}
    }
    \hspace{-0.18cm}
    \subfigure[GT]{
    \begin{minipage}[t]{0.18\linewidth}
    \includegraphics[width=1.35in]{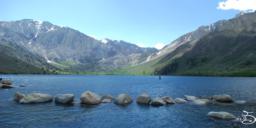}\\
    \vspace{-0.4cm}
    \includegraphics[width=1.35in]{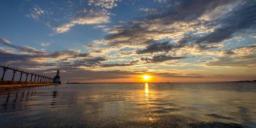}\\
     \vspace{-0.4cm}
    \includegraphics[width=1.35in]{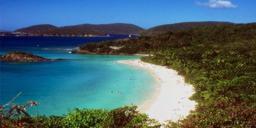}\\
     \vspace{-0.3cm}
    \end{minipage}
    }
\end{center}
\vspace{-0.3cm}
 \caption{Three exemplary results for the image restoring according to the the original sketches. The transition boundary of Pix2Pix is not smooth enough, which makes the final images look separate. The NSIO and the BDIE can achieve acceptable transitions, but the generated parts lack reality. Our results not only achieves a smooth transition from the left to the right, but the semantic consistency is satisfactory.}
 %\vspace{-0.2cm}
\label{fig:competing results}
\end{figure*}

\begin{figure*}[t]
\begin{center}
    \subfigure[The inputs]{
    \begin{minipage}[t]{0.18\linewidth}
     \includegraphics[width=1.35in]{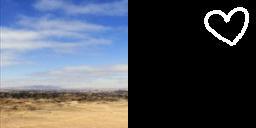}\\
      \vspace{-0.4cm}
    \includegraphics[width=1.35in]{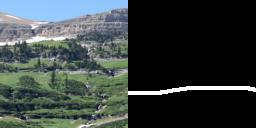}\\
      \vspace{-0.4cm}
    \includegraphics[width=1.35in]{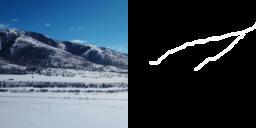}\\
    \vspace{-0.3cm}
    \end{minipage}
    }
   \hspace{-0.18cm}
    \subfigure[Pix2Pix~\cite{I2I_2017}]{
    \begin{minipage}[t]{0.18\linewidth}
    \includegraphics[width=1.35in]{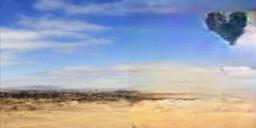}\\
    \vspace{-0.4cm}
    \includegraphics[width=1.35in]{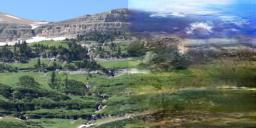}\\
    \vspace{-0.4cm}
    \includegraphics[width=1.35in]{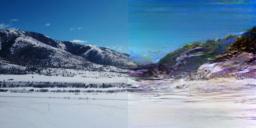}\\
     \vspace{-0.3cm}
    \end{minipage}}
    \hspace{-0.06cm}
    \subfigure[NSIO~\cite{yzx}]{
    \begin{minipage}[t]{0.18\linewidth}
    \includegraphics[width=1.35in]{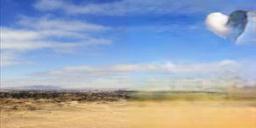}\\
    \vspace{-0.4cm}
    \includegraphics[width=1.35in]{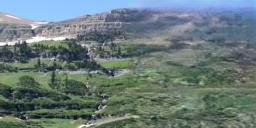}\\
    \vspace{-0.4cm}
    \includegraphics[width=1.35in]{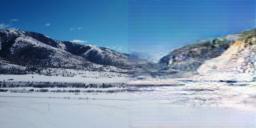}\\
     \vspace{-0.3cm}
    \end{minipage}
    }
    \hspace{-0.18cm}
    \subfigure[BDIE~\cite{boundless}]{
    \begin{minipage}[t]{0.181\linewidth}
     \includegraphics[width=1.35in]{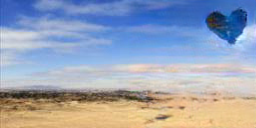}\\
     \vspace{-0.4cm}
    \includegraphics[width=1.35in]{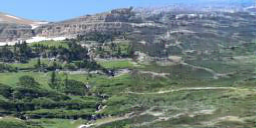}\\
    \vspace{-0.4cm}
    \includegraphics[width=1.35in]{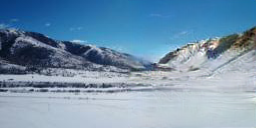}\\
     \vspace{-0.3cm}
    \end{minipage}
    }
    \hspace{-0.18cm}
    \subfigure[Ours]{
    \begin{minipage}[t]{0.18\linewidth}
     \includegraphics[width=1.35in]{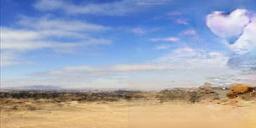}\\
     \vspace{-0.4cm}
    \includegraphics[width=1.35in]{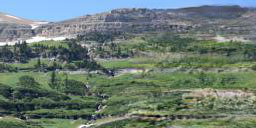}\\
    \vspace{-0.4cm}
    \includegraphics[width=1.35in]{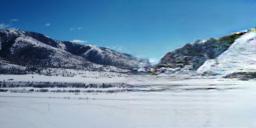}\\
     \vspace{-0.3cm}
    \end{minipage}
    }
\end{center}
\vspace{-0.3cm}
 \caption{Results for sketch-guided outpainting. All the comparison methods suffer from the sudden transition around the boundary or the unreasonable pixel filling for the desired shape, as a consequence, they all fail to produce satisfactory results. While our model could achieve a smooth transition from the original input to the synthesized right half and preserve the semantic consistency of the entire image very well.}
 %\vspace{-0.2cm}
\label{visual_competing_free}
\end{figure*}

%Besides the commonly used reconstruction loss and the adversarial loss, we propose a sketch-consistence loss to recover the high frequency information in images. Let $\hat{S}$ be the sketch of the synthesized image, which can be obtained by feeding $\hat{I}$ through the edge detection module, our sketch-consistence loss is defined as follow:
%\begin{equation}
%    \mathcal{L}_{s} = ||\hat{S} - S||^2_2.
%\end{equation}
The model is trained via a combination of the adversarial loss and our proposed sketch alignment loss. Our discriminator loss comprises of the global discriminator loss $\mathcal{L}_d$, and the local discriminator loss $\mathcal{L}^{'}_d$, which can be obtained according to the Eq.~\ref{d_loss}, by feeding the inputs to corresponding discriminator module. In summary, the full discriminator loss reads:
\begin{equation}
    \mathcal{L}_D = \mathcal{L}_d + \mathcal{L}^{'}_d,
\end{equation}
and the generator loss is formulated as:
\begin{equation}
    \mathcal{L}_G = \lambda_r\mathcal{L}_1 +\lambda_s\mathcal{L}_s+ \lambda_a[\alpha\mathcal{L}_g + (1-\alpha)\mathcal{L}^{'}_g],
\end{equation}
where $\lambda_r, \lambda_s,\lambda_a$ and $\alpha$ are the trade-off weights.

 In our practice, we find directly training the network could successfully restore the images according to the original sketches, but can not perform well on free-form outpainting. The reason may stem from the sketch overfitting. \yx{Since the model is trained using only the original sketches, the deep impression for these original sketches pose obstacles to generalize to the free-form sketches. To remedy this issue, we propose to augment the sketch pattern by a random sketch masking mechanism. } Our designed sketch augmentation is only applied on the right half sketch,  since the outpainting model conditions on the right sketch.  For current right half sketch in  training stage:
 \begin{itemize}
     \item  make the sketch unchanged with probability 0.4,
     \item mask a randomly selected patch whose scale ranges from 48$\times$48 to 64$\times$128, in the top part and the bottom part with probability 0.2 and 0.4, respectively.
 \end{itemize}
 The sketches in bottom part are paid more attention, because we find the sketches in bottom are richer and more complex than the ones in top part. 
 \begin{figure*}[t]
\setlength{\abovecaptionskip}{0pt} 
\setlength{\belowcaptionskip}{0pt} 
\begin{center}
    \subfigure{
    \begin{minipage}[t]{0.18\linewidth}
     \includegraphics[width=1.35in]{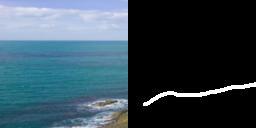}\\
     \vspace{-0.4cm}
    \includegraphics[width=1.35in]{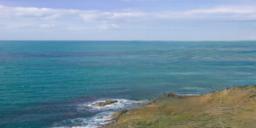}\\
    \vspace{-0.4cm}
    \includegraphics[width=1.35in]{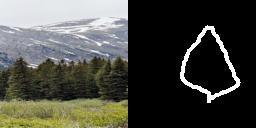}\\
    \vspace{-0.4cm}
    \includegraphics[width=1.35in]{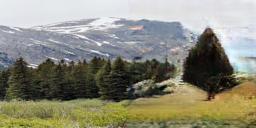}
  
    \end{minipage}
    }
    \hspace{-0.06cm}
    \subfigure{\begin{minipage}[t]{0.18\linewidth}
    \includegraphics[width=1.35in]{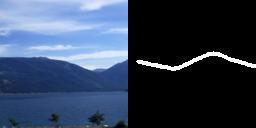}\\
    \vspace{-0.4cm}
    \includegraphics[width=1.35in]{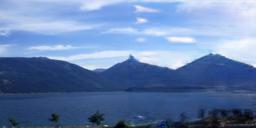}\\
    \vspace{-0.4cm}
    \includegraphics[width=1.35in]{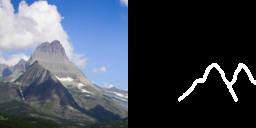}\\
    \vspace{-0.4cm}
    \includegraphics[width=1.35in]{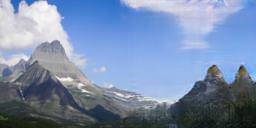}
    \end{minipage}}
    \hspace{-0.06cm}
    \subfigure{
    \begin{minipage}[t]{0.18\linewidth}
   \includegraphics[width=1.35in]{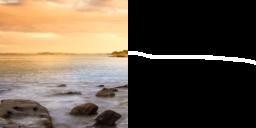}\\
    \vspace{-0.4cm}
    \includegraphics[width=1.35in]{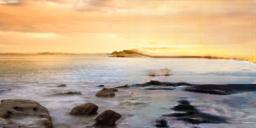}\\
    \vspace{-0.4cm}
     \includegraphics[width=1.35in]{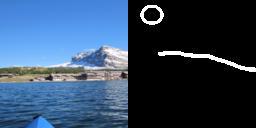}\\
    \vspace{-0.4cm}
    \includegraphics[width=1.35in]{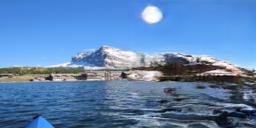}
    \end{minipage}
    }
    \hspace{-0.18cm}
    \subfigure{
    \begin{minipage}[t]{0.18\linewidth}
   \includegraphics[width=1.35in]{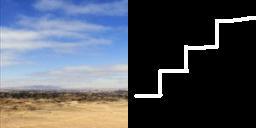}\\
   \vspace{-0.4cm}
    \includegraphics[width=1.35in]{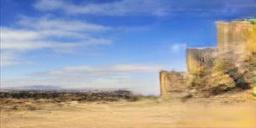}\\
    \vspace{-0.4cm}
    \includegraphics[width=1.35in]{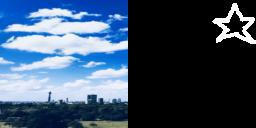}\\
    \vspace{-0.4cm}
    \includegraphics[width=1.35in]{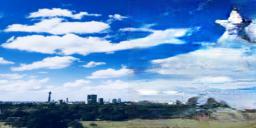}
    \end{minipage}
    }
    \hspace{-0.18cm}
    \subfigure{
    \begin{minipage}[t]{0.18\linewidth}
     \includegraphics[width=1.35in]{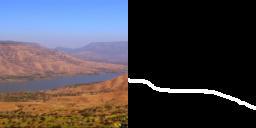}\\
     \vspace{-0.4cm}
    \includegraphics[width=1.35in]{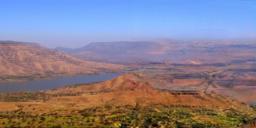}\\
    \vspace{-0.4cm}
   \includegraphics[width=1.35in]{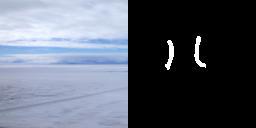}\\
   \vspace{-0.4cm}
    \includegraphics[width=1.35in]{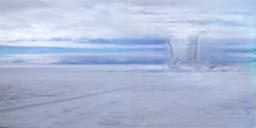}
   
    \end{minipage}
    }
\end{center}
\caption{Several outpainting results for free-form sketches. Our method successfully generates natural images with diverse freestyle sketches.}
\label{fig:final exhibition}
\end{figure*}

\section{Experimental Results}
\label{experiment}
\subsection{Experiment setup}

\subsubsection{Datasets} 
The NS6K dataset~\cite{yzx}, which consists of 6,040 natural scenery images, is employed to evaluate our method, and the data split follows Yang's setting~\cite{yzx}. However, we find most of the sketches from HED~\cite{HED} in NS6K are simple, and there are many images similar to each other. To test our method under more practical situations, we first pick up 4,040 images from NS6K by filtering out 2,000 similar images, and then collect 4,075 images from the `Google images' by utilizing eighteen scenery keywords as index including \yx{Alps, Himanayas Nepal, La Digue, Moraine Lake, Napali Coast, Preikestolen, The Bay of Fundy, Arches park, Grassland, Niagara, Mount Cook, National Forest Park, Olympic National Park, Qinghai Lake, South Island, Sunset, Yellowstone Park, Pattaya Beach. In total, we collect 5014 images using these keywords as indexes. We then filter out some low-quality images as well as similar samples, then 4075 images are collected. Then image number for each keyword is shown in Fig.~\ref{sample_dis}. The collected 4075 images and the picked 4040 samples from NS6K form a more diverse and complex dataset called NS8K, which contains 8115 natural scenery images.} Of these, 1,500 images are for testing while the rest is taken as training data.
%Therefore, we construct a more diverse and complex dataset based on the NS6K. We first pick up the top 2,000 image pairs according to their similarity, which is calculated based on the features from pre-trained VGG16 network~\cite{vgg16}. For an image pair, only one of them is retained, and additional 4,075 images is further collected from the internet by utilizing eighteen scenery keywords as indexing including Alps, Arches Park, La Digue, etc. The collected images and the retained images from NS6K form a new dataset called NS8K, which contains 8115 natural scenery images. Of these, 1,500 images are for testing while the rest is taken as training data.

\begin{figure*}[t]
\begin{center}
    \subfigure{
    \begin{minipage}[t]{0.9\linewidth}
    \includegraphics[width=6.5in]{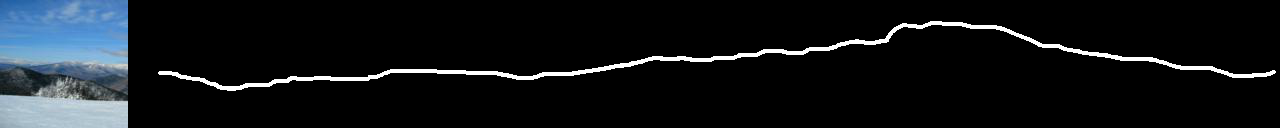}\\
    %\vspace{-0.3cm}
    \includegraphics[width=6.5in]{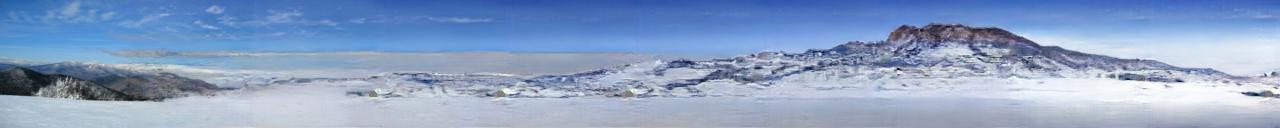}\\
    %\vspace{-0.3cm}
    \includegraphics[width=6.5in]{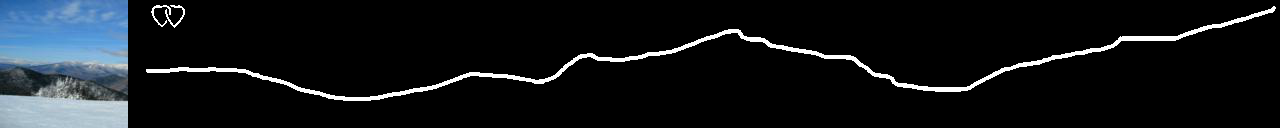}\\
    %\vspace{-0.3cm}
    \includegraphics[width=6.5in]{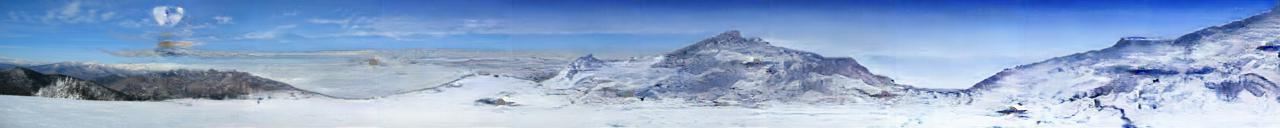}\\
    \end{minipage}
    }
\end{center}
\vspace{-0.6cm}
 \caption{The results for multi-step prediction. The longer results could be synthesized by taking current outpainting for next-step prediction.}
 %\vspace{-0.2cm}
\label{multi_step}
\end{figure*}

\subsubsection{Implement Details}
Following Yu \etal~\cite{Yu0}, the sketch in this work is obtained using the HED edge detector~\cite{HED}, we first extract the edge map, and set the values above 0.6 as 1 to get the binary sketch. 
To maintain better consistency, the pre-trained HED detector, whose parameters are frozen during training, is also used for edge map detection in our sketch alignment module. 
Our network synthesizes $128\times 256$ image by providing an image and a guiding sketch with shape 128$\times$128 as inputs. For the generator, the image concatenated with the sketch is first passed through a gated convolution layer~\cite{Yu0}, while the raw position channels are encoded by a convolution layer. Two feature maps are channel-wise concatenated and fed forward through several gated convolution layers, to get a tensor with shape 4$\times$4$\times$512.  
%compresses the inputs including raw image, sketch and position channels to a feature map with shape 4$\times$ 4$\times$ 512, which is reshaped as 2048$\times$4, i.e. $N=2048, K=4$.
The encoder LSTM~\cite{LSTM} collaborating with the decoder LSTM produce another tensor with shape $4\times4\times512$, which couples with the left one to rebuild the complete image with shape 128$\times$256 by a series of gated deconvolution layers~\cite{Yu0}. The feature dimension of the hidden state for both LSTMs is set as 2048. We choose the gated structure because of its superior in dealing with the binary sketch~\cite{boundless,Yu0}. 
Our network is implemented using the tensorflow platform~\cite{tensorflow} and trained on 2 NVIDIA GTX 1080Ti GPUs. The parameters of the generator and discriminators are jointly updated using the Adam optimizer~\cite{adam} with batch size 30. The weight $\lambda_r$ for reconstruction loss is set as 0.998, while the adversarial loss weight $\lambda_a$ is set as 0.002. Hyperparameters $\lambda_s$, $\alpha$ and $\lambda_w$ are fixed as 1, 0.9 and 10, respectively. The training iteration is up to 800 epochs and starts with learning rate 0.0001, which is discounted by 0.1 after 200 epochs. 

\yx{Table~\ref{architecture} shows the detailed architecture of our generator, where the first column shows the layer type, the following two columns explain the input and output shapes, respectively, and the parameter configurations are exhibited in the last column. Specifically, the `G-Conv’/`G-Deconv’ indicates the gated convolution/deconvolution [2], respectively. `G-Resblock$\times\#$’ means `$\#$’ G-Resblocks are stacked, where the G-Resblock is the Resblock with the convolution replaced by the G-Conv. And the `Sketch' Encoder and the `Position Encoder' are shallow neural networks which are composed by several convolution \& pooling operations. The LSTM encoder/decoder are both LSTM cells. For the shapes shown in ‘input’ and ‘output’ columns,  they are organized as height$\times$width$\times$channel. }

\begin{figure*}[t]
%\centering
%\setlength{\abovecaptionskip}{0pt} 
%\setlength{\belowcaptionskip}{0pt} 
\begin{center}
\subfigure[Baseline]{
    \begin{minipage}[t]{0.18\linewidth}
        \centering
        \includegraphics[width=1.35in]{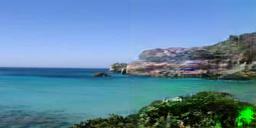}\\
        \vspace{0.05cm}
        \includegraphics[width=1.35in]{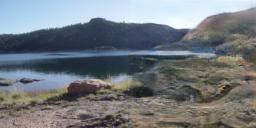}\\
        \vspace{0.05cm}
         \includegraphics[width=1.35in]{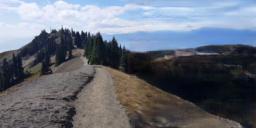}\\
        \vspace{0.2cm}
        \label{aba_re:a}
    \end{minipage}
    }
    \hspace{-0.18cm}
\subfigure[+ PCs]{
    \begin{minipage}[t]{0.18\linewidth}
        \centering
        \includegraphics[width=1.35in]{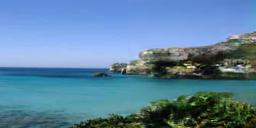}\\
          \vspace{0.05cm}
        \includegraphics[width=1.35in]{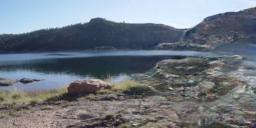}\\
          \vspace{0.05cm}
         \includegraphics[width=1.35in]{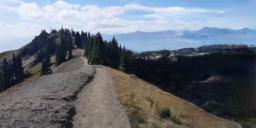}\\
        \label{aba_re:b}
          \vspace{0.2cm}
    \end{minipage}
    }
    \hspace{-0.18cm}
\subfigure[+ CSC]{
    \begin{minipage}[t]{0.18\linewidth}
        \centering
        \includegraphics[width=1.35in]{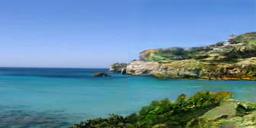}\\
         \vspace{0.05cm}
        \includegraphics[width=1.35in]{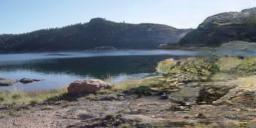}\\
         \vspace{0.05cm}
         \includegraphics[width=1.35in]{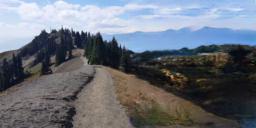}\\
        \label{aba_re:c}
          \vspace{0.2cm}
    \end{minipage}
    }
    \hspace{-0.18cm}
\subfigure[+ SAL]{
    \begin{minipage}[t]{0.18\linewidth}
        \centering
         \includegraphics[width=1.35in]{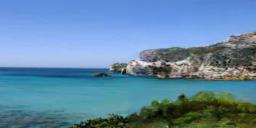}\\
         \vspace{0.05cm}
        \includegraphics[width=1.35in]{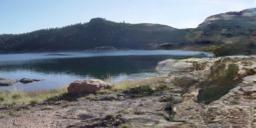}\\
         \vspace{0.05cm}
         \includegraphics[width=1.35in]{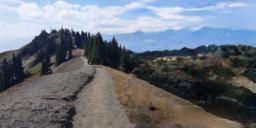}\\
        \label{aba_re:d}
          \vspace{0.2cm}
    \end{minipage}
    }
    \hspace{-0.18cm}
\subfigure[GT]{
    \begin{minipage}[t]{0.18\linewidth}
        \centering
         \includegraphics[width=1.35in]{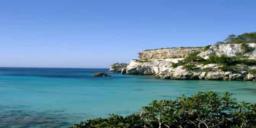}\\
          \vspace{0.05cm}
        \includegraphics[width=1.35in]{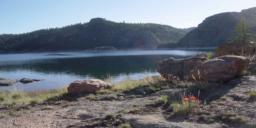}\\
          \vspace{0.05cm}
         \includegraphics[width=1.35in]{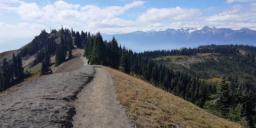}\\
        \label{aba_re:e}
          \vspace{0.2cm}
    \end{minipage}
    }
    \vspace{-0.2cm}
\caption{Visual ablation comparison on image rebuilding, where the PCs, CSC, SAL and GT represent the position channels, conditional skip connection, sketch alignment loss and groundtruth, respectively.}
%\text{I. Image Rebuilding}
\label{abaltion_restore}
\end{center}
\end{figure*}

\begin{figure*}[t]
\begin{center}
\subfigure[Inputs]{
    \begin{minipage}[t]{0.18\linewidth}
        \centering
        \includegraphics[width=1.35in]{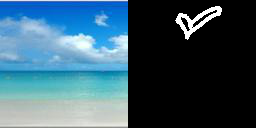}\\
          \vspace{0.05cm}
         \includegraphics[width=1.35in]{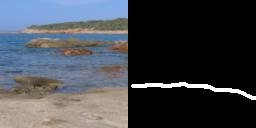}\\
           \vspace{0.05cm}
          \includegraphics[width=1.35in]{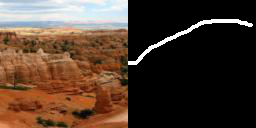}\\
        \vspace{0.2cm}
        \label{aba_free:a}
    \end{minipage}
    }
\subfigure[Baseline]{
    \begin{minipage}[t]{0.18\linewidth}
        \centering
        \includegraphics[width=1.35in]{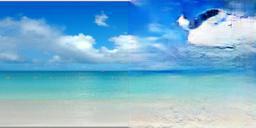}\\
          \vspace{0.05cm}
        \includegraphics[width=1.35in]{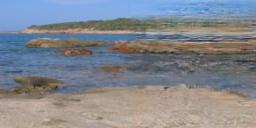}\\
          \vspace{0.05cm}
         \includegraphics[width=1.35in]{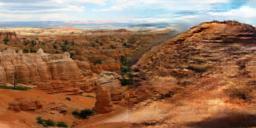}\\
         \vspace{0.2cm}
        \label{aba_free:b}
    \end{minipage}
    }
\subfigure[+ PCs]{
    \begin{minipage}[t]{0.18\linewidth}
        \centering
        \includegraphics[width=1.35in]{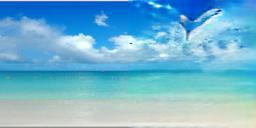}\\
         \vspace{0.05cm}
        \includegraphics[width=1.35in]{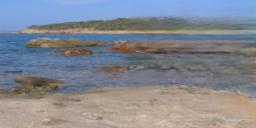}\\
         \vspace{0.05cm}
         \includegraphics[width=1.35in]{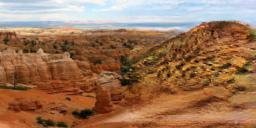}\\
         \vspace{0.2cm}
        \label{aba_free:c}
    \end{minipage}
    }
\subfigure[+ CSC]{
    \begin{minipage}[t]{0.18\linewidth}
        \centering
         \includegraphics[width=1.35in]{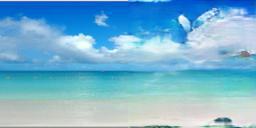}\\
          \vspace{0.05cm}
        \includegraphics[width=1.35in]{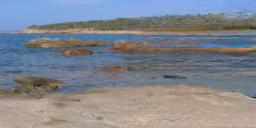}\\
          \vspace{0.05cm}
         \includegraphics[width=1.35in]{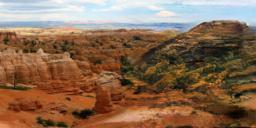}\\
         \vspace{0.2cm}
        \label{aba_free:d}
    \end{minipage}
    }
\subfigure[+ SAL]{
    \begin{minipage}[t]{0.18\linewidth}
        \centering
         \includegraphics[width=1.35in]{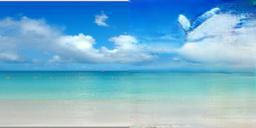}\\
           \vspace{0.05cm}
        \includegraphics[width=1.35in]{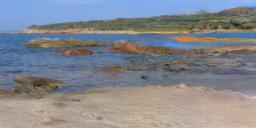}\\
          \vspace{0.05cm}
         \includegraphics[width=1.35in]{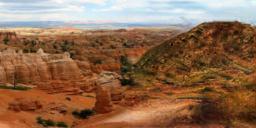}\\
         \vspace{0.2cm}
        \label{aba_free:e}
    \end{minipage}
    }
%\text{II. Free-form outpainting}
\vspace{-0.2cm}
\caption{Visual ablation comparison on free-form outpainting, the results get improved when a new module is equipped.}
\label{ablation_free}
\end{center}
\end{figure*}

%\vspace{-0.5cm}
\subsubsection{Evaluation Metric}
Three criteria are used to evaluate the proposed method, \ie the Fr$\acute{\rm e}$chet Inception Distance (FID)~\cite{FID}, the Inception Score (IS)~\cite{IS} and the Mean Satisfaction Degree (MSD). To conduct an objective comparison, we feed the original sketches from the test dataset to rebuild the corresponding images, and the FID and IS can be obtained according to the synthesized images and the groundtruth data:
\begin{align}
    &\text{IS} = \exp(\mathbb{E}_{\hat{I}\sim \mathbb{P}_f}D_{KL}(p(y|\hat{I})||p(y))),\\
    &\text{FID} = ||\mu_r - \mu_f||^2 + Tr(\Sigma_r + \Sigma_f -2(\Sigma_r\Sigma_f)^{1/2}),
\end{align}
where $\text{IS}$ is defined based on the KL-divergence between the classification distribution of the fake sample and the mean probability on each class, while the $\text{FID}$ first employs the Inception-V3 network~\cite{inceptionv3} to extract the 2048-d features and then computes the statistics distance to evaluate the generation model, the $\mu_f \text{and } \Sigma_f$ are the mean vector and the covariance matrix of the fake features, $Tr$ is the trace operation. 

As for the free-form outpainting, we can not obtain objective performance since there is no grountruth available. Therefore, 
%to evaluate whether the synthesized images meet the guiding sketches from users, and the overall quality of freestyle image outpainting, 
we employ a subjective metric, \ie the Mean Satisfactory Degree (MSD), to evaluate the quality of free-form outpainting. First, we randomly select 300 images from the test dataset and replace their original right sketches with manually drawn free-form ones, there are 77 different types of sketches in total. Then, 20 volunteers are invited to label the satisfaction degree of each synthesized sample as three levels: 0-poor, 1-ordinary and 2-good. The mean value of all labels on test images is taken as the mean satisfaction degree (MSD). Comparing to image restoring from the original sketches, using free-form sketches for outpainting is closer to the practical situation, since the right half image is usually not available. Therefore, the MSD is more important for performance evaluation.

\subsection{Quantitative Comparisons}
%\textbf{Performance} 
%To evaluate the synthesis quality of the proposed model, we feed the original sketches from the test dataset to rebuild the corresponding images, the Fr$\acute{\rm e}$chet Inception Distance (FID)~\cite{FID} and the Inception Score(IS)~\cite{IS} are calculated for objective performance comparing. 
Table~\ref{NS6k_results} and Table~\ref{NS8k_results} show the results of our method and three competing methods on NS6K and NS8K dataset, respectively.
%\footnote{all of these competing methods are modified to conduct sketch-guided image outpainting task.} including two outpainting methods NSIO~\cite{yzx} and BDIE~\cite{boundless} 
The comparison methods include two outpainting methods NSIO~\cite{yzx}, BDIE~\cite{boundless}, and a classic image to image translation work, Pix2Pix~\cite{I2I_2017}. All comparison methods conduct the same data augmentation including randomly cropping and flipping,  and are trained by 1,500 epoch iterations. The Pix2Pix~\cite{I2I_2017} is trained by the loss functions in~\cite{yzx}, the right half of the input image is masked and channel-wise concatenated with the sketch to translate to the original image.  For NSIO~\cite{yzx}, the left sketch and the right half sketch are used in the same way as our method. For BDIE~\cite{I2I_2017}, the sketch is directly channel-wise concatenated with the masked input to synthesize the full image.
%Both BDIE[21] and NSIO[25] employ the gate convolution~\cite{Yu0}.

%the results between the groundtruth images by two widely used criteria Inception Score(IS)~\cite{IS} and Fr$\acute{\rm e}$chet Inception Distance (FID)~\cite{FID}. 
\setlength{\tabcolsep}{1.5mm}{
\begin{table}[t]
\setlength{\abovecaptionskip}{0pt}%    
\setlength{\belowcaptionskip}{2pt}%
\begin{center}
\begin{tabular}{|c|cccc|}
\hline
Method & Pix2Pix ~\cite{I2I_2017} & NSIO ~\cite{yzx}&BDIE ~\cite{boundless}&ours\\
\hline
FID${\color{red}\downarrow}$ &21.197  &13.17 &13.424&\textbf{10.998}\\
IS${\color{red} \uparrow}$ & 2.783&2.887& 2.899 & \textbf{2.920}\\
\hline\hline
MSD${\color{red} \uparrow}$ &0.472 & 0.544 & 0.777 & \textbf{1.027}\\
\hline
\end{tabular}
\end{center}
%\vspace{-0.2cm}
\caption{The results of four methods on NS6K under evaluation criteria IS (the higher the better), FID (the lower the better) and MSD (the higher the better).}
\label{NS6k_results}
\vspace{-0.3cm}
\end{table}
}
\setlength{\tabcolsep}{1.5mm}{
\begin{table}[t]
\setlength{\abovecaptionskip}{0pt}%    
\setlength{\belowcaptionskip}{2pt}%
\begin{center}
\begin{tabular}{|c|cccc|}
\hline
Method & Pix2Pix ~\cite{I2I_2017} & NSIO ~\cite{yzx}&BDIE ~\cite{boundless}&Ours\\
\hline
FID${\color{red}\downarrow}$ & 18.327  &11.153 &10.891&\textbf{10.390}\\
IS${\color{red} \uparrow}$ & 3.013 & 3.254&3.276& \textbf{3.321}\\
%\hdashline{1-5}
\hline
\hline
MSD${\color{red} \uparrow}$ &0.615 & 0.706 & 0.725 & \textbf{1.031}\\
\hline
\end{tabular}
\end{center}
\caption{The performance of four methods on the NS8K dataset.}
\label{NS8k_results}
%\vspace{-0.4cm}
\end{table}
}

As shown in Table~\ref{NS6k_results},  the FID of our method can reach 10.998, which is much better than all competing methods. On the free-form outpainting,  the superiority of our method is more obvious. Our method could achieve 1.027 MSD, while the competitive methods' are only 0.777 at best. From Table~\ref{NS8k_results}, our method is still much more outstanding for the free-form outpainting on NS8K dataset. As shown in Table~\ref{NS8k_results}, the competing method, BDIE~\cite{boundless}, achieves comparable performance on the image rebuilding task. For example, our FID is 10.390 while the FID of BDIE~\cite{boundless} could reach 10.891. However, using free-form sketches for outpainting is closer to the practical situation since the right half image is usually not available. Although the competing methods could achieve the acceptable performance on image rebuilding according to the original sketches, they perform much worse on the free-form outpainting. The MSD of BDIE~\cite{boundless} is only 0.725 on NS8K dataset, and the other two comparison methods are even much worse. In contrast, our method performs much better on the free-form outpainting and could achieve 1.031 MSD, which surpasses the comparison methods by a large margin.  
%The similar observation can be obtained from the results on NS8K (Table~\ref{NS8k_results}).
%Additionally, to evaluate whether the synthesized images meet the guiding sketches from users and the overall quality of freestyle image outpainting, we randomly select 300 images from the test dataset and replace their original right sketches with manually drawn free-form ones, there are 77 different types of sketch in total. The satisfaction degree for a generated image is manually labeled as three levels by five volunteers: 0-poor, 1-ordinary and 2-good. The mean value of all labels on test images are taken as the mean satisfaction degree (MSD), which is taken for subjective evaluation since there is no groundtruth available.  
From Table~\ref{NS6k_results} and Table~\ref{NS8k_results}, the existing outpainting methods could deal with the original sketches but fail to generalize to the more practical situation, \ie the free-form outpainting. While our proposed method not only achieves the best performance on image rebuilding according to the original sketches but could harvest more satisfactory results for the free-style image outpainting on both datasets, which validates the effectiveness of our proposed approach. 

%The improvement of our method over others seems not obvious on NS8K according to IS and FID. The reason is that we only take the ground-truth sketches extracted from testing images for evaluation. Only in this way, we could acquire quantitative results. However, using free-form sketches for outpainting is closer to the practical situation since the right half image is usually not available. With a free-form-based evaluation metrics (i.e., MSD), our method performs much better (as shown in Table 1, 2), which well demonstrates the generalization ability of our method over others. 
%Exemplary results shown in Fig.~\ref{fig:competing results} and Fig.~\ref{fig:competing results with original sketch} intuitively show the superiority of our method. 
\subsection{Qualitative Results}
%Fig.~\ref{fig:competing results} and Fig.~\ref{visual_competing_free} show the qualitative comparison of the four methods on the image rebuilding and free-form outpainting, respectively. 
\yx{Fig.~\ref{fig:competing results} subsequently shows three image rebuilding examples of four methods, Pix2Pix~\cite{I2I_2017}, NSIO~\cite{yzx}, BDIE~\cite{boundless} and Ours, where the left half image is the input, and the guiding sketches are directly extracted from the right half of the ground-truth images. In Fig.~\ref{visual_competing_free}, the comparison for free-form outpainting is presented, where the first column exhibits the input images and the guiding sketches and the following second to the fourth columns subsequently show the results of methods, Pix2Pix~\cite{I2I_2017}, NSIO~\cite{yzx}, BDIE~\cite{boundless}, and our model. }
As shown in Fig.~\ref{fig:competing results}, the Pix2Pix~\cite{I2I_2017} could not achieve the smooth transition around the boundary, since there is no module or loss designed to stitch the boundary in its architecture. As for the NSIO~\cite{yzx} and BDIE~\cite{boundless}, the boundary between the original image and the synthesized part is relatively smooth, however, the results still suffer from the lack of textural details and disharmonious pixels. While our method could not only achieve the smooth transition from the left to the right half but could synthesize results with more textural details. The superiority of our method is more obvious on the free-form outpainting, as shown in Fig.~\ref{visual_competing_free}.  All the competing methods could not ensure the semantic consistency and the smooth boundary and fail to fill reasonable pixels for the free-style sketches. Even for the simple sketch like a single line (the second row in Fig.~\ref{visual_competing_free}), the comparison methods could not make a success, which reveals the poor generalization of these methods and the challenges of the free-form outpainting task. While our proposed network could successfully predict the reasonable pixels for the free-style sketches by the learned positional prior knowledge, which helps the network synthesize much more natural results. 

%Fig.~\ref{first page exhibition} and 
\yx{Fig.~\ref{fig:final exhibition} exhibits several groups of outpainting results with many diverse free-form sketches, where the first and the third rows show the inputs as well as the guiding sketches, and their corresponding outputs are exhibited in the second and the fourth rows.} As shown in these two figures, our method could not only successfully conduct outpainting for the sketches similar to the training data but could well generalize to the unseen sketches like circle shape, heart shape. Thanks to the learned prior positional knowledge, even though the provided sketch does not correspond to semantically meaningful content as shown in the fourth column Fig.~\ref{fig:final exhibition}, our approach could also fill the reasonable pixels for the sketch and produce relatively semantically consistent content. From Fig.~\ref{first page exhibition} and Fig.~\ref{fig:final exhibition}, it can be intuitively seen that the proposed method can produce natural outpainting results for diverse manually drawn sketches. Even though most of the provided sketches are missed in the training set, our model can still generate realistic images and preserve the semantic consistency well. Besides, our method could also synthesize longer images by taking current output as the input for next prediction. \yx{In Fig.~\ref{multi_step}, we exhibit two examples for 9-step prediction, where the upper rows show the inputs, and the counterpart outputs are displayed in the next rows. The new content are iteratively produced by taking the output in previous step as input.}

\yxnew{From Fig.~\ref{fig:competing results}-\ref{ablation_free}, we could see that our method could synthesize satisfactory results for the image rebuilding but fail to produce texture-rich contents according to the free-form sketches. The essential reason for such an observation is the lack of training data for free-form sketches.  Since there are no available training samples for the free style sketches, the network is trained using the original sketches to restore the full image. Consequently, the model would have a deep ``impression'' for the original sketches, and could produce more details by directly invoking the memory of the pixels and the sketch patterns. For the free-form outpainting, the situation becomes much challenging due to the unseen sketch patterns, the network needs to ``imagine'' the reasonable contents  for the free-style sketches according to the memory. Although we have made some attempts to ease the inferring for free-form outpainting including introducing the position channels and conducting augmentation to extend the sketch patterns, the image quality around the free-form sketches is still not that perfect. The gap between the training sketches and the free-form sketches poses an important challenge for the sketch-guided outpainting. In the future, we would continue to explore this meaningful task and develop approaches to produce texture-rich results.}

\subsection{Ablation Study}
To validate the effectiveness of each component in our system, we conduct ablation study on NS6K to verify their respective contributions, results are reported in Table~\ref{ablation study}. Our baseline only employs the sketch and the Gated Convolution ~\cite{Yu0} to conduct outpainting.
%$SL, PC, RM$ means seaming loss, position channels and random mask mechanism respectively. 
From Table~\ref{ablation study}, the model with four parts simultaneously utilized achieves the best MSD, and when a new mechanism is equipped, the MSD gets improved, which validates the contribution of each component.
\setlength{\tabcolsep}{2mm}{
\begin{table}[t]
\setlength{\abovecaptionskip}{0pt}%    
\setlength{\belowcaptionskip}{-10pt}%
\begin{center}
\begin{tabular}{|ccccc|ccc|}
\hline
 &RSM & PCs & CSC & SAL &IS${\color{red} \uparrow}$ &FID${\color{red} \downarrow}$ &MSD${\color{red} \uparrow}$\\
\hline
baseline &  & & &   &2.889 & 12.87 & 0.587\\
baseline &\checkmark & & & &2.882 &12.27 &0.661\\
%baseline &\checkmark  &  & &2.867 &12.19 &0.736\\
baseline &\checkmark &\checkmark & &  &2.873 &11.526 &0.747\\
baseline &\checkmark &\checkmark & \checkmark &\ &2.883 &11.451 &0.879\\
baseline &\checkmark &\checkmark & \checkmark &\checkmark  &\textbf{2.920} &\textbf{10.998} &\textbf{1.027}\\
\hline
\end{tabular}
\end{center}
\caption{The contribution of each part. RSM, PCs, CSC and SAL indicate random sketch masking, position channels, conditional skip connection and sketch alignment loss, respectively.}
\label{ablation study}
\end{table}
}
\begin{figure}[t]
    \centering
    \includegraphics[height=2.6in, width=3.5in]{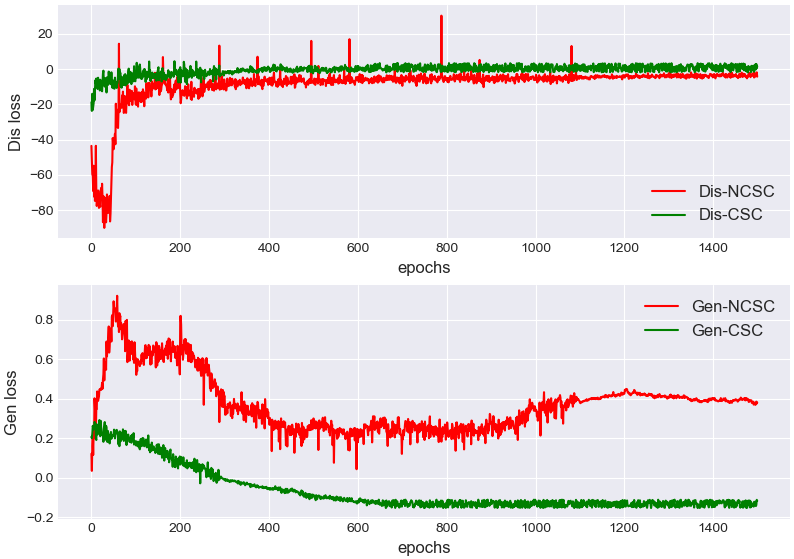}
    %\vspace{-0.2cm}
    \caption{The loss tendencies in our training procedure. The curves with `CSC' means the training loss with our conditional skip connection, while `NCSC' indicates the loss without the CSC module.}
    %\vspace{-0.6cm}
    \label{converge}
\end{figure}

Although the random sketch masking strategy and position channels cause a slight performance drop according to the Inception Score, they could both boost the FID and the MSD, therefore these two strategies play important roles in our system. The FID and IS in Table~\ref{ablation study} show that the conditional skip connection makes a little contribution to the image rebuilding, however, it could effectively improve the free-style outpainting.  \yx{Fig.~\ref{csc_abalation} intuitively shows the contribution of the CSC module to our system, where the first column exhibits the input images and the guiding sketches, the results with/without the CSC module are shown in second and third columns, respectively, and the last column displays the original images. From Fig.~\ref{csc_abalation} (b), there exists some distortion in the outpainting from the generator without CSC module, as a result, the results fail to match the guiding sketches. By highlighting the expecting sketches in decoding stage using the CSC module, the generator could synthesize the results that are exactly consistent with the guiding sketches. }

What's more, we find the CSC can also make a faster convergence for the network training. Fig.~\ref{converge} shows the tendencies of the average discriminator and generator losses in every training epoch. Our network without the CSC requires around 1,500 epochs training, while once the CSC module is equipped, the network converges after about 700 epochs, which remarkably speeds the network training. From Fig.~\ref{converge}, our CSC module could also make the training steadier. Furthermore, when the sketch alignment loss is equipped, our method achieves the best performance with FID and MSD scores of 10.998 and 1.027, respectively.

%Exemplary results for ablation study and more result exhibition can be found in our supplemental material.
\yx{Fig.~\ref{abaltion_restore} and Fig.~\ref{ablation_free} show the visual ablation on image rebuilding and free-form outpainting to intuitively validate the contribution of each component. From left to right, the modules are subsequently stacked to synthesize the results. For example, the “+PCs” column shows the outpainting generated by the baseline with position channels,  where the “+CSC” column shows the results from the baseline with position channels and conditional skip connection simultaneously equipped.}. As shown in these two figures, the baseline method produces some abrupt pixels which make the overall image not authentic enough, and the synthesized results suffer from the lack of textural details, especially for the free-form outpainting.  By introducing the position channels, the model could predict the reasonable pixels with the learned positional relation between the pixels and the specific positions. The sketch alignment loss imposes the generator to restore the high-frequency information, consequently, when the sketch alignment loss is equipped, the boundary of different semantic regions is clearer, and more details could be observed from Fig.~\ref{abaltion_restore} and Fig.~\ref{ablation_free}. From Fig.~\ref{abaltion_restore}, the contribution of the conditional skip connection is not obvious for image rebuilding, but it is important for the free-style outpainting on preventing the desired sketch from distorting, as shown in top row in Fig.~\ref{ablation_free}.

%The baseline method with only sketch used can not get satisfactory results. From the top rows in Fig~\ref{abaltion_restore} and Fig~\ref{ablation_free}, we can find that there are some abrupt pixels in the results. By introducing the position channels, the model could predict the reasonable pixels for the new content. The contribution of the CSC is not that clear in image rebuilding,
%but it is important for the free-style outpainting on preventing the desired sketch from distorting, as shown in top row example in Fig.~\ref{ablation_free}. When the sketch alignment loss is equipped, the boundary of different semantic regions is clearer, and more details can be observed.  

\section{Conclusion and Future Work}
\label{conclusion}
We have presented the first solution for the under explored sketch-guided image outpainting problem, which is a meaningful yet challenging task. The developed framework allows users to guide the outpainting results by free-style sketches. Our encoder compresses inputs to hidden features, and the decoder integrates the hidden features and the guiding information to build the desired image. Specifically, two position channels are introduced for reasonable pixel filling, and a conditional skip connection is proposed to make the results spatial consistent with the guiding sketch. To restore the high-frequency details, we design the sketch alignment loss to further boost the outpainting quality. In addition, we contribute a more complex and diverse scenery image dataset NS8K for further image outpainting study. Experiments on two benchmarks demonstrate the effectiveness and the ability of our model on sketch-guided image outpainting. Although the proposed method could outperform all existing image outpainting models,  the results of free-style sketches still suffer from the lack of textural details. In our future work, we would develop model to compensate more textural details for the free-style outpainting.

% or


\iffalse
\appendices
\section{Proof of the First Zonklar Equation}
Appendix one text goes here.

\section{}
Appendix two text goes here.

\section*{Acknowledgment}

The authors would like to thank...
\fi

\ifCLASSOPTIONcaptionsoff
  \newpage
\fi

\end{document}